\documentclass[a4paper]{article} 
\usepackage[submission]{aaai23}  
\usepackage{times}  
\usepackage{helvet}  
\usepackage{courier}  
\usepackage[hyphens]{url}  
\usepackage{graphicx} 
\usepackage{natbib}  
\usepackage{caption} 
\usepackage{algorithm}
\usepackage{algorithmic}

\usepackage[T1]{fontenc}
\usepackage[font=small,labelfont=bf,tableposition=top]{caption}

\usepackage{makecell}

\usepackage{tabularx}

\usepackage{tabularx}

\newtheorem{theorem}{Theorem}

\newtheorem{assumption}{Assumption}

\DeclareCaptionLabelFormat{andtable}{#1~#2  \&  \tablename~\thetable}

\newtheorem{lemma}{Lemma}
\newtheorem{corollary}{Corollary}

\usepackage[T1]{fontenc}
\usepackage[utf8]{inputenc}
\usepackage{babel}
\usepackage[font=small,labelfont=bf]{caption}
\usepackage{subcaption}
\usepackage{graphicx}
\usepackage{capt-of}
\usepackage{xcolor,colortbl}
\usepackage{footmisc}
\usepackage{soul}

\definecolor{LightCyan}{rgb}{0.88,1,1}

\usepackage[utf8]{inputenc} 
\usepackage[T1]{fontenc}    
\usepackage{nameref}
\usepackage{hyperref}       
\usepackage{url}            
\usepackage{booktabs}       
\usepackage{amsfonts}       
\usepackage{nicefrac}       
\usepackage{microtype}      
\usepackage{xcolor}         
\usepackage{multirow}

\usepackage{newfloat}
\usepackage{listings}
\DeclareCaptionStyle{ruled}{labelfont=normalfont,labelsep=colon,strut=off} 
\floatstyle{ruled}
\newfloat{listing}{tb}{lst}{}
\floatname{listing}{Listing}
\title{Efficient Distribution Similarity Identification in Clustered Federated Learning via Principal Angles Between Client Data Subspaces}
 
 \author{%
Saeed Vahidian$^{1*}$ \quad Mahdi Morafah$^{1*}$ \quad Weijia Wang$^1$ \quad Vyacheslav Kungurtsev$^2$ \\
\textbf{Chen Chen}$^3$ \quad \textbf{Mubarak Shah}$^3$ \quad \textbf{Bill Lin}$^1$ \\
$^1$UC San Diego \quad $^2$Czech Technical University \quad $^3$ UCF\\
}

\usepackage{bibentry}

\begin{document}

\maketitle

\begin{abstract}
\vspace{-2mm}
Clustered federated learning (FL) has been shown to produce promising results by grouping clients into clusters.
This is especially effective in scenarios where separate groups of clients have significant differences in the distributions of their local data. Existing clustered FL algorithms are essentially trying to group together clients with similar distributions so that clients in the same cluster can leverage each other's data to better perform federated learning. However, prior clustered FL algorithms attempt to learn these distribution similarities indirectly during training, which can be quite time consuming as many rounds of federated learning may be required until the formation of clusters is stabilized. In this paper, we propose a new approach to federated learning that directly aims to efficiently identify distribution similarities among clients by analyzing the principal angles between the client data subspaces. Each client applies a truncated singular value decomposition (SVD) step on its local data in a single-shot manner to derive a small set of principal vectors, which provides a signature that succinctly captures the main characteristics of the underlying distribution.
This small set of principal vectors is provided to the server so that the server can directly identify distribution similarities among the clients to form clusters.
This is achieved by comparing the similarities of the principal angles between the client data subspaces spanned by those principal vectors. The approach provides a simple, yet effective clustered FL framework that addresses a broad range of data heterogeneity issues beyond simpler forms of Non-IIDness like label skews. Our clustered FL approach also enables convergence guarantees for non-convex objectives.
\end{abstract}

\vspace{-6mm}
\section{Introduction}
  \vspace{-1mm}
Federated Learning (FL)~\cite{mcmahan2017federated} 
enables a set of clients to collaboratively learn a shared prediction model without sharing their local data.
Some FL approaches aim to train a common global model for all clients~\cite{mcmahan2017communication, fedprox-smith-2020, FedNova-2020, scaffold-2020, ChenChen-cvpr-2021}.
However,
in many FL applications where there may be data heterogeneity among clients, a single relevant global model may not exist. Alternatively, personalized FL approaches have been studied.
One approach
is to first train a global model and then allow each client to fine-tune it via a few rounds
of stochastic gradient descent (SGD)~\cite{fallah2020personalized, liang2020think}.
Another approach is for each client to jointly train a global model as well as a local model, and then interpolate them to derive a personalized model~\cite{mahdavi2020-APFL, Mansour-federated-2020}. 
In the former case, the approach often fails to derive a model that generalizes well to the local distributions of each client.
In the latter case, when local distributions and the average distribution are far apart, the approach often degenerates to every client learning only on its own 
local data. Recently, clustered FL~\cite{Ghosh-federated-2020, sattler-clustered-fl-2021, Mansour-federated-2020} has been proposed to allow the grouping of clients into clusters so that clients belonging to the same cluster can share the same optimal model.
Clustered FL has been shown to produce significantly better results, especially 
when separate groups of clients have significant differences in the distributions of their local data. This possibly due to distinct learning tasks or the mixture of distributions of the local data considered, not necessarily limited to simpler forms of data heterogeneity such as label skews from otherwise the same dataset.

Essentially, what prior clustered FL algorithms are trying to do is to group together clients with similar distributions so that clients in the same cluster can leverage each other's data to perform federated learning more effectively. 
Previous clustered FL algorithms attempt to \emph{learn} these distribution similarities \emph{indirectly} when clients learn the cluster to which they should belong as well as the cluster model during training.
For example, the clustered FL approach presented in~\cite{sattler-clustered-fl-2021} alternately estimates the cluster identities of clients and optimizes the cluster model parameters via SGD.



Unfortunately, the prior clustered FL approaches have the following \emph{challenges} which in turn limits their applicability in real-world problems. $1)$ Since previous clustered FL training algorithms start with randomly initialized cluster models that are inherently noisy, the overall training process can be quite time consuming as many rounds of federated learning may be required until the formation of clusters is stabilized.
$2)$ Approaches like IFCA~\cite{Ghosh-federated-2020} assumes a pre-defined number of clusters, but requiring the number of clusters to be fixed \emph{a priori}, regardless of the differences in the actual data distributions or learning tasks among the clients, could lead to poor performance for many clients. $3)$ In each iteration, all cluster models have to be downloaded by the active clients in that round, which can be very costly in communications. $4)$ Both of the approaches i.e., those that train a common global model for all clients and personalized approaches including IFCA lack the flexibility to trade off between \emph{personalization} and \emph{globalization}. The above-mentioned drawbacks of the prior works, naturally lead to the following important question.\emph{ How a server can realize clustered FL efficiently by grouping the clients into clusters in a one-shot manner without requiring the number of clusters to be known apriori, but with substantially less communication cost?}  In this work, we propose a novel algorithm, \underline{P}rincipal \underline{A}ngles analysis for \underline{C}lustered \underline{F}ederated \underline{L}earning (PACFL), to address the above-mentioned challenges of clustered FL.

\noindent\textbf{Our contributions.}
We propose a new algorithm, PACFL, for federated learning that \emph{directly} aims to efficiently identify distribution similarities among clients by analyzing the \emph{principal angles between the client data subspaces}.
Each client wishing to join the federation
applies a truncated SVD step on its local data in a \emph{one-shot} manner to derive a \emph{small set of principal vectors}, which form the principal bases of the underlying data. These principal bases provide a \emph{signature} that succinctly captures the main characteristics of the underlying distribution.
The client then provides this small set of principal vectors to the server so that the server can directly identify distribution similarities among the clients to form clusters.
The privacy of data is preserved since no client data is ever sent to the server but a few $(2$-$5)$ principal vectors out of $\approx500$. Thus, the clients data cannot be reconstructed from those $(2$-$5)$ number of left singular vectors. However, in privacy sensitive setups to provide extra protection and prevent any information leakage from clients to server, mechanisms like the ones presented in~\cite{Privacy-Preserving-FL-2017}, or encryption mechanism or differential privacy method that achieves this end can be employed.


On the server side, it efficiently identifies distribution similarities among clients by comparing the principal angles between the client data subspaces spanned by the provided principal vectors -- 
the greater the difference in data heterogeneity between two clients,
the more orthogonal their subspaces.
Unlike prior clustered FL approaches, which require time consuming iterative learning of the clusters and substantial communication costs, our approach provides a simple yet effective clustered FL framework that addresses a broad range of data heterogeneity issues beyond simpler forms of Non-IIDness like label skews.
Clients can immediately collaborate with other clients in the same cluster from the get go.

Our novel PACFL approach has the flexibility to trade off between \emph{personalization} and \emph{globalization}. PACFL can naturally span the spectrum of identifying IID data distribution scenarios in which all clients should share training within only 1 cluster, to the other end of the spectrum where clients have extremely Non-IID data distributions in which each client would be best trained on just its own local data (i.e., each client becomes its cluster).

Our framework also naturally provides an elegant approach to handle newcomer clients unseen at training time by matching them with a cluster model that the client can further personalized with local training.
Realistically, new clients may arrive to the federation after the distributed training procedure.
In our framework, the newcomer client simply provides its principal vectors to the server, and the server identifies via angle similarity analysis which existing cluster model would be most suitable, or the server can inform the client that it should train on its own local data to form a new cluster if the client's data distribution is not sufficiently similar to the distributions of the existing clusters. On the other hand, it is generally unclear how prior personalized or clustered FL algorithms can be extended to provide newcomer clients with similar capabilities.


Finally, we provide a convergence analysis of 
PACFL in the supplementary material.


\vspace{-3mm}
\section{Clustered Federated Learning}


\vspace{-1mm}
\subsection{Preliminaries}
\label{Preliminaries}
\vspace{-1mm}
\noindent \textbf{Principal angles between two subspaces.}  Let ${\mathcal{U}} = \rm{span}\{{\bf{u}}_1, . . . , {\bf{u}}_p\}$ and ${\mathcal{W}} = \rm{span}\{{\bf{w}}_1, . . . , {\bf{w}}_q\}$ be $p$ and $q$-dimensional subspaces of $\bf{R^n}$ where $\{{\bf{u}}_1, . . . , {\bf{u}}_p\}$ and $\{{\bf{w}}_1, . . . , {\bf{w}}_q\}$ are orthonormal, with $1 \le p \le q$. There exists a sequence of $p$ angles $0 \le \Theta_1 \le \Theta_2 \le ...\le \Theta_p \le \pi/2$ called the principal angles, which are defined as
\vspace{-2mm}
\begin{equation}
\small
\Theta\left({\mathcal{U}}, {\mathcal{W}} \right)=\min\limits_{{\bf{u}} \in {\mathcal{U}}, {\bf{w}} \in {\mathcal{W}}}  \arccos\left( \frac{\left| {\bf{u}}^T{\bf{w}} \right| }{\left\| {\bf{u}} \right\| \left\| {\bf{w}}\right\| }
 \right)
\end{equation}
\vspace{-4mm}

\noindent where $\left\| .\right\|$ is the induced norm. The smallest principal angle is $\Theta_1 \left( {\bf{u}}_1,{\bf{w}}_1\right)$ with vectors ${\bf{u}}_{1}$ and ${\bf{w}}_{1}$ being the corresponding principal vectors. The principal angle distance is a metric for measuring the distance between subspaces~\cite{principal-angle-2013}. Additional background is presented in the supplementary material.

\vspace{-3mm}
\begin{figure}[!ht]
    \centering
\includegraphics[width=55mm, height=30mm]{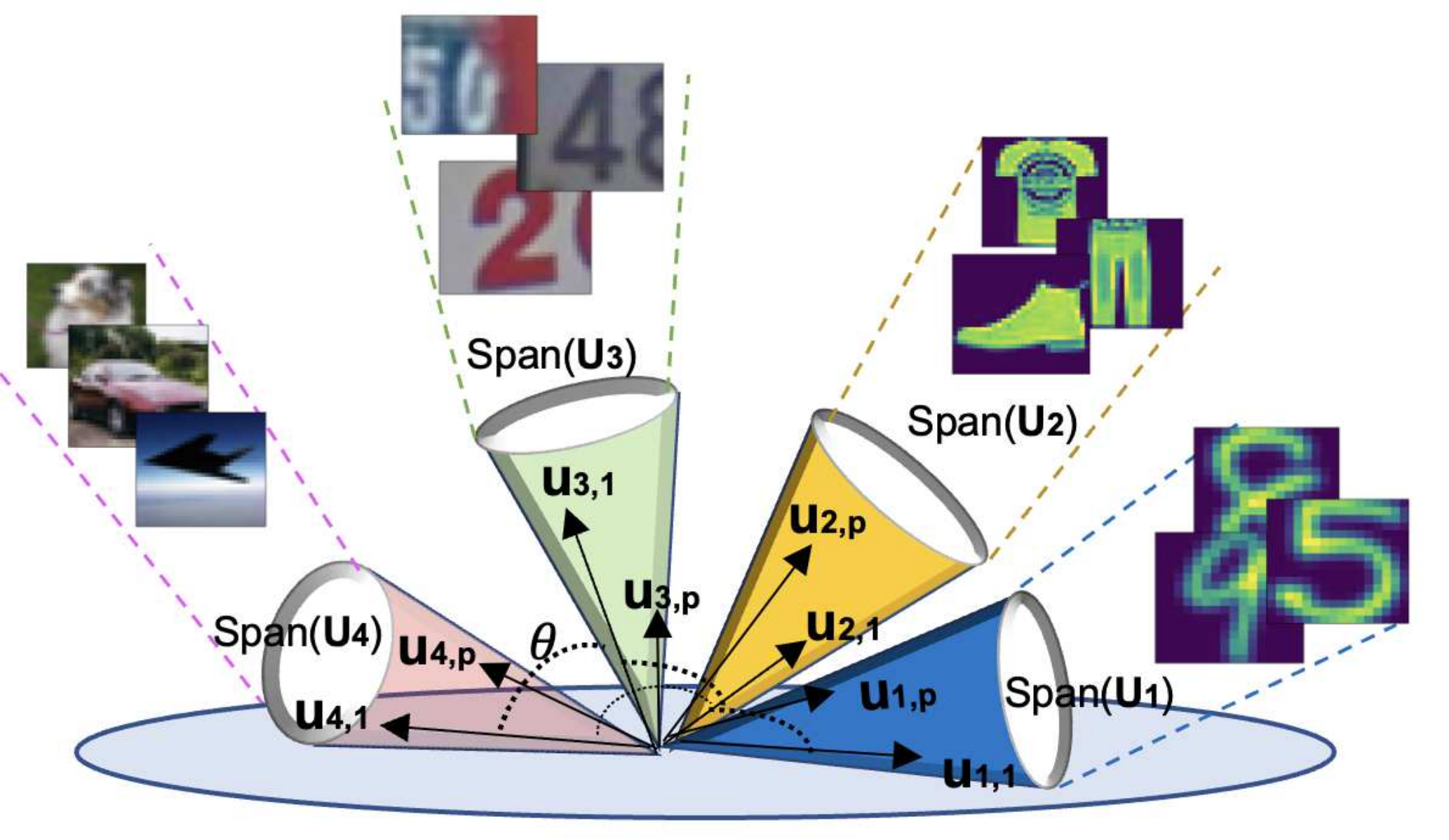}
\vspace{-0.5mm}
    \qquad

\scalebox{0.72}{
		    	\begin{tabular}{|c|c|c|c|c|c|}

 \hline
 Dataset & CIFAR-10 & SVHN  & FMNIST & USPS  \\
		
\hline
	
CIFAR-10& 0 (0) & 6.13 (12.3) & 45.79 (91.6) & 66.26 (132.5) \\
		\hline
SVHN & 6.13 (12.3)& 0 (0) & 43.42 (86.8) & 64.86 (129.7)\\
		\hline
FMNIST & 45.79 (91.6)& 43.42 (86.8) & 0 (0) & 43.36 (86.7)\\
		\hline
USPS & 66.26 (132.5) & 64.86 (129.7)& 43.36 (86.7)& 0 (0)\\
		\hline
  \end{tabular} \label{cc}
     }
    \captionlistentry[table]{A table beside a figure}
    \captionsetup{labelformat=andtable}
 
 \vspace{-2mm}
    \caption{Fig.~1: \footnotesize{There must be a translation protocol enabling the server to understand similarity/dissimilarity of the clients data without sharing data. This 2D figure intuitively demonstrates how the principal angle between the client data subspaces captures the statistical heterogeneity. Fig. 1 Shows the subspaces spanned by the $\mathbf{U}_p$s of four different datasets (left to right: CIFAR-10, SVHN, FMNIST, and USPS). As can be seen the principal angle between the subspaces of CIFAR-10 and SVHN is smaller than that of CIFAR-10 and USPS.} Table 1: \footnotesize{This table demonstrates how distribution similarities among different datasets can be accurately estimated by the principal angles between the datasets subspaces. This table shows the proximity matrix of four datasets whose entries are the exact principal angles between every pairs of these datasets' subspaces. Entries are $x (y)$, where $x$ and $y$ are respectively the smallest principal angle, and the summation over the principal angles between two datasets. $p$ in $\mathbf{U}_p$ is 2.}}
    
    \vspace{-6mm}
  \end{figure}

\vspace{-1mm}
\subsection{How Principal Angles Can Capture the Similarity Between Data/Features}

The \emph{cosine similarity} is a distance metric between vectors that is known to be more tractable and interpretable than the alternatives while exhibiting high precision clustering properties, see, e.g.~\cite{qian2004similarity}. The idea is to note that for any two vectors $x$ and $y$, by the dot product calculation $x\cdot y=\|x\|\|y\|\cos\theta$, we see that inverting the operation to solve for $\theta$ yields the angle between two vectors as emanating from the origin, which is a scale-invariant indication of their alignment. It presents a natural geometric understanding of the proportional volume of the embedded space that lies between the two vectors. Finally, by choosing to cluster using the data rather than the model, the variance of each SGD sample declines resulting in smoother training. By contrast, clustering by model parameters has the effect of increasing the bias of the clients' models to be closer to each other.

In order to obtain a computationally tractable small set of vectors to represent the data features, we propose to apply truncated SVD on each dataset. We take a small set of principal vectors, which form 
the principal bases of the underlying data distribution. Truncated SVD is known to yield a good quality balance between computational expense and representative quality of representative subspace methods~\cite{talwalkar2013large}. Assume there are $K$ number of datasets. We propose to apply truncated SVD (detailed in the supplementary material) on these data matrices, $\mathbf{D}_k, k=1,...,K$, whose columns are the input features of each dataset. Further, let $ \mathbf{U}^k_p=\left[  {\mathbf{u}_1, \mathbf{u}_2,..., \mathbf{u}_p} \right]$, $\left(p \ll \rm{rank}\left(\mathbf{D}_k\right)\right)$ be the $p$ most significant left singular vectors for dataset $k$. We constitute the proximity matrix $\mathbf{A}$ as in Eq.~\ref{adj1} whose entries are the smallest principle angle between the pairs of $\mathbf{U}^k_p$ or as in Eq.~\ref{adj2} whose entries are the summation over the angle in between of the corresponding $\mathbf{u}$ vectors (in identical order) in each pair within $\mathbf{U}^k_p$, where $\mathbf{tr}\left(.\right)$ is the trace operator, and


 \vspace{-3mm}
\begin{equation}
    \mathbf{A}_{i,j}=\Theta_1 \left(  \mathbf{U}^i_p, \mathbf{U}^j_p\right),~~i,j=1,...,K
    \label{adj1}
\end{equation}
\vspace{-4mm}
\begin{equation}
\mathbf{A}_{i,j}=\mathbf{tr}\left(\arccos \left(  \mathbf{U}^{i~T }_p* \mathbf{U}^j_p\right)\right),~~i,j=1,...,K
    \label{adj2}
\end{equation}
\vspace{-3mm}

The smaller the entry of $\mathbf{A}_{i,j}$ is, the more similar datasets $i$ and $j$ are~\footnote{It is noteworthy that in practice both of these equations work accurately. However, theoretically and rigorously speaking, when the number of the principal vectors, $p$, is bigger than 1, it can happen that one of the principal vectors of client $k$ yields a small angle with its corresponding one for client $k'$ while the other principal vectors of client $k$ yield big angle with their corresponding ones for client $k'$. With that in mind, Eq.~\ref{adj2} is a more rigorous measure and it always truly captures the similarity between the client data subspaces. 
}. Before we proceed further, through some experiments on benchmark datasets, we highlight how the proposed method perfectly distinguishes different datasets based on their hidden data distribution by inspecting the angle between their data subspaces spanned by their first $p$ left singular vectors. For a visual illustration of the result, we refer to Fig.~1. As can be seen the principal angle between the subspaces of CIFAR-10 and SVHN is smaller than that of CIFAR-10 and USPS. Table~1 shows the exact principal angles between every pairs of these datasets' subspaces. The entries of this table is presented as $x(y)$, where $x$ is the smallest principal angle between two datasets obtained from Eq.~\ref{adj1}, and $y$ is the summation over the principal angles between two datasets obtained from Eq.~\ref{adj2}. Table~1 reveals that the similarity and dissimilarity of the four different datasets have been accurately captured by the proposed method.  We will provide more examples in the supplementary material and will show that the similarity/dissimilarity being captured by the proposed method is consistent with well-known distance measures between two distributions including Bhattacharyya Distance (BD), Maximum Mean Discrepancy (MMD)~\cite{MMD2012}, and Kullback–Leibler (KL) distance~\cite{KL-Gaussian}. \vspace{-3mm}

\begin{algorithm}[th]
{\fontsize{8pt}{8pt}\selectfont
\caption{ PACFL}
\label{alg:PACFL-2}
\begin{algorithmic}[1]
\STATE \textbf{Require:} {Number of available clients $N$, sampling rate $R\in(0,1]$, clustering threshold $\beta$}
\STATE \textbf{Server:} Initialize the server model with $\theta_g^0$.
\FOR {each round $t$ = 1, 2,...} {
\STATE $m \leftarrow {\rm{max}}(R \cdot N,1)$
\STATE $\mathcal{S}_t \leftarrow \{k_1,...,k_n\}$ \COMMENT {\scriptsize{set of $n$ available clients}}
\FOR  {each client $k \in {{\mathcal{S}}_t}$ \rm{\textbf{in parallel}}}
{ 
\IF{$t=1$ \COMMENT {\scriptsize{\textcolor{blue}{It is done in one-shot} }}}  
{\STATE client $k$ sends $\mathbf{U}^k_p$ to the server
\STATE $\mathbf{U} = [\mathbf{U}^{k_1}_p, ..., \mathbf{U}^{k_n}_p]$
\STATE $\mathbf{A} \leftarrow$ server forms $\mathbf{A}$ based on Eq.~\ref{adj1} or Eq.~\ref{adj2} 
\STATE $\{C_1,...,C_Z\} = \rm{HC}(\mathbf{A}, \beta)$
\STATE $\theta^0_{g,z} \leftarrow \theta_g^0$ \COMMENT {\scriptsize{initializing all clusters with $\theta_g^0$}}
}
\ELSIF{$k$ is a new arriving client}{
\STATE client $k$ sends $\mathbf{U}^k_p$ to the server
\STATE $\mathbf{A}$, $\mathbf{U}$ = $\rm{PME}(\mathbf{A}$, $\mathbf{U}$, $\mathbf{U}^k_p)$ \COMMENT {\scriptsize{Alg.~\ref{alg:Func}}}
\STATE $\{C_1,...,C_Z\} \leftarrow \rm{HC}(\mathbf{A}, \beta)$  \COMMENT {\scriptsize{Update the clusters, determine the cluster ID of new client $k$} }
\STATE client $k$ receives the corresponding cluster model $\theta^t_{g,z} $ from the server
}
\ELSE{
\STATE client $k$ sends its cluster ID to the server and receives the corresponding cluster model $\theta^t_{g,z} $ from the server
    }
\ENDIF
\STATE $\theta^{t+1}_{k,z}\leftarrow {\rm{ClientUpdate}}(k; \theta^t_{g,z} )$: by SGD training 
}
\ENDFOR
\FOR{$z=1:Z$ ~~~\COMMENT {\scriptsize{$Z$ is the number of formed clusters}}}{  
\STATE $\theta^{t+1}_{g,z}=\sum_{k \in C_{z} }{|D_{k}|\theta^{t+1}_{k,z}}  /\sum_{k \in C_{z} }{|D_{k}|}$ \COMMENT{\scriptsize{model averaging for each cluster}}
}
\ENDFOR
}
\ENDFOR
 \end{algorithmic}
 }
 \vspace{-0.7mm}
\end{algorithm}

\vspace{-8mm}
\begin{algorithm}[th]
{\fontsize{8pt}{8pt}\selectfont
\caption{\footnotesize{Proximity Matrix Extension (PME)}}
\label{alg:Func}
\begin{algorithmic}[1]
\STATE \textbf{Input: }{$\mathbf{A}_{old}$ \COMMENT{\scriptsize{ $M \times M$ proximity matrix formed by $M$ number of seen clients}}}     
\STATE \textbf{Input: }{$\mathbf{U}_{old} = [U_p^1,...,U_p^M]$ \COMMENT{\scriptsize{ Set of first $p$ significant singular vectors of the $M$ seen clients}} }
\STATE \textbf{Input: }{$\mathbf{U}_{new} = [U_p^1,...,U_p^B]$ \COMMENT{\scriptsize{The set of first $p$ significant singular vectors of the $B$ new  clients}} }

\STATE \textbf{Output: }{$\mathbf{A}_{extended}$ \COMMENT{\scriptsize{The extended $(M+B)\times(M+B)$ proximity matrix}}}
\STATE \textbf{Output: }{$\mathbf{U}_{extended}$} 

\FUNCTION {PME($\mathbf{A}_{old}$, $\mathbf{U}_{old}$, $\mathbf{U}_{new}$)}
{
\STATE $\mathbf{A}_{extended} \leftarrow \mathbf{[0]}_{(M+B)\times(M+B)}$
\STATE $U_{extended} \leftarrow [\mathbf{U}_{old}, \mathbf{U}_{new}]$ 
\STATE $\mathbf{A}_{extended}[1:M, 1:M] = \mathbf{A}_{old}[:, :]$
\STATE $\mathbf{A}_{extended}[M:M+B, M:M+B]$ \COMMENT {can be calculated based on Eq.~\ref{adj1} or Eq.~\ref{adj2}}
\STATE \textbf{Return} $\mathbf{A}_{extended}$, $\mathbf{U}_{extended}$
  }
\ENDFUNCTION
\end{algorithmic}
}
\end{algorithm}
\normalsize

\subsection{Overview of PACFL}
\vspace{-1mm}
\label{overview}
In this section, we begin by presenting our PACFL framework. 
The proposed approach, PACFL, is described in Algorithm~\ref{alg:PACFL-2}. We first turn our attention to clustering clients data in a federated network. The proposed method is one-shot clustering and can be used as a simple pre-processing stage to characterize personalized federated learning to achieve superior performance relative to the recent iterative approach for clustered FL proposed in~\cite{Ghosh-federated-2020}. Before federation, each available client, $k$, performs truncated SVD on its own data matrix, $\mathbf{D}_k$~\footnote{Considering a client owns M data samples, each including N features, we assumed that the M data samples are organized as the columns of a matrix $D_k \in R^{N \times M}$.}, and sends the $p$ most significant left singular vectors $ \mathbf{U}_p$, as their data \emph{signature} to the central server.  Next, the server obtains the proximity matrix $\mathbf{A}$ as in Eq.~\ref{adj1} or Eq.~\ref{adj2} where $K=|{\mathcal{S}}_t|$, and ${\mathcal{S}}_t$ is the set of available clients. When the number of clusters is unknown, for forming disjoint clusters, the server can employ agglomerative hierarchical clustering (HC)~\cite{day1984efficient} on the proximity matrix $\mathbf{A}$. \emph{For more details on HC, please see the supplementary material.} Hence, the cluster ID of clients is determined.



For training, the algorithm starts with a single initial model parameters $\theta^0_g$. In the first iteration of PACFL a random subset of available clients ${\mathcal{S}}_t \subseteq [N]$, $|{\mathcal{S}}_t|=n$ is selected by the server and the server broadcasts $\theta^0_g$ to all clients. The clients start training on their local data and perform some steps of stochastic gradient descent (SGD) updates, and get the updated model. The clients will only need to send their cluster membership ID and model parameters back to the central server. After receiving the model and cluster ID memberships from
all the participating clients, the server then collects all the parameter updates from clients whose cluster ID are the same and conducts model averaging within each cluster. It is noteworthy that in Algorithm~\ref{alg:PACFL-2}, $\beta$ stands for the Euclidean distance between two clusters and is a parameter in HC.


\textbf{Desirable properties of PACFL.} Unlike prior work on clustered federated learning~\cite{Ghosh-federated-2020, sattler-clustered-fl-2021}, PACFL has much greater flexibility in the following sense. First, from a \emph{practical} perspective, one of the desirable properties of PACFL is that it can handle partial participation of clients. In addition, PACFL does not require to know in advance whether certain clients are available for participation in the federation. Clients can join and leave the network abruptly. In our proposed approach, the new clients that join the federation just need to send their data signature to the server and the server can easily determine the cluster IDs of the new clients by constituting a new proximity matrix without altering the cluster IDs of the other clients. In PACFL, the prior information about the availability of certain clients is not required.

Second, PACFL can form the best fitting number of clusters, if a fixed number of clusters is not specified. However, in IFCA~\cite{Ghosh-federated-2020}, the number of clusters has to be known apriori. Third, one-shot client clustering can be placed by PACFL for the available clients before the federation and the prior information about the availability and the number of certain clients is not required. In contrast, IFCA constructs the clusters iteratively by alternating between cluster identification estimation and loss function minimization which is costly in communication. 

Fourth, PACFL does not add significant additional computational overhead to the FedAvg baseline algorithm as it
only requires running one-shot HC clustering before training. With that in mind, the computational complexity of the PACFL algorithm is the same as that of FedAvg plus the computational complexity of HC in one-shot ($(O(N^2))$  where $N$ is the total number of clients).

Fifth, in case of either a certain and a fixed number of clients are not available at the initial stage or clients join and leave the network abruptly, clustering by PACFL can easily be applied in a few stages as outlined in Algorithm~\ref{alg:Func}. Each new clients that become available for the federation, sends the signature of its data to the server and the server aggregates the signature of existing clients and the new ones as in $U_{extended}$ (Algorithm~~\ref{alg:Func}). Next, the server obtains the proximity matrix $\mathbf{A}$ as in Eq. 3 or Eq. 4 where all the new clients included. \emph{By keeping the same distance threshold} as before, the cluster ID of the new clients are determined without changing the cluster ID of the old clients.



Sixth, we should note that we tried some other clustering methods including graph clustering methods~\cite{Boyd-2015-clustering, Yasmin-iccasp-2021} for PACFL and we noticed that the clustering algorithm does not play a crucial role in PACFL. As long as the clustering algorithm itself does not require the number of clusters to be known in advance, it can be applied in PACFL.

Seventh, the server cannot make use of the well known distribution similarity measures such as BD, MMD~\cite{MMD2012}, and KL~\cite{KL-Gaussian} to group the clients into clusters due to the privacy constraints as they require accessing the data or important moments of the data distributions. As shown in Fig.~1, and Table~1 and also as will be shown in the Experiments Section, the proposed approach presents a simple and alternative solution to the above-mentioned measures in FL setups.

We also provide a convergence analysis of the method in the supplementary material. The framework we use is from~\cite{haddadpour2019convergence}, which considers nonconvex learning with Non-IID data. Indeed unlike other works we can obtain guarantees for nonconvex objectives, as appropriate for deep learning, because the clustering is performed on the \emph{data} and not the parameters, thus no longer suffering from associated issues of multi-modality (multiple separate local minima). We shall see that it recovers the state-of-the-art (SOTA) convergence rate and performance guarantees for Non-IID data.

\normalsize

\vspace{-4mm}
\section{Experiments}
\label{exp:section}
\vspace{-1mm}


\begin{figure*}[htbp]
  \begin{minipage}{\textwidth}
    \centering
  \includegraphics[width=.18\pdfpagewidth]{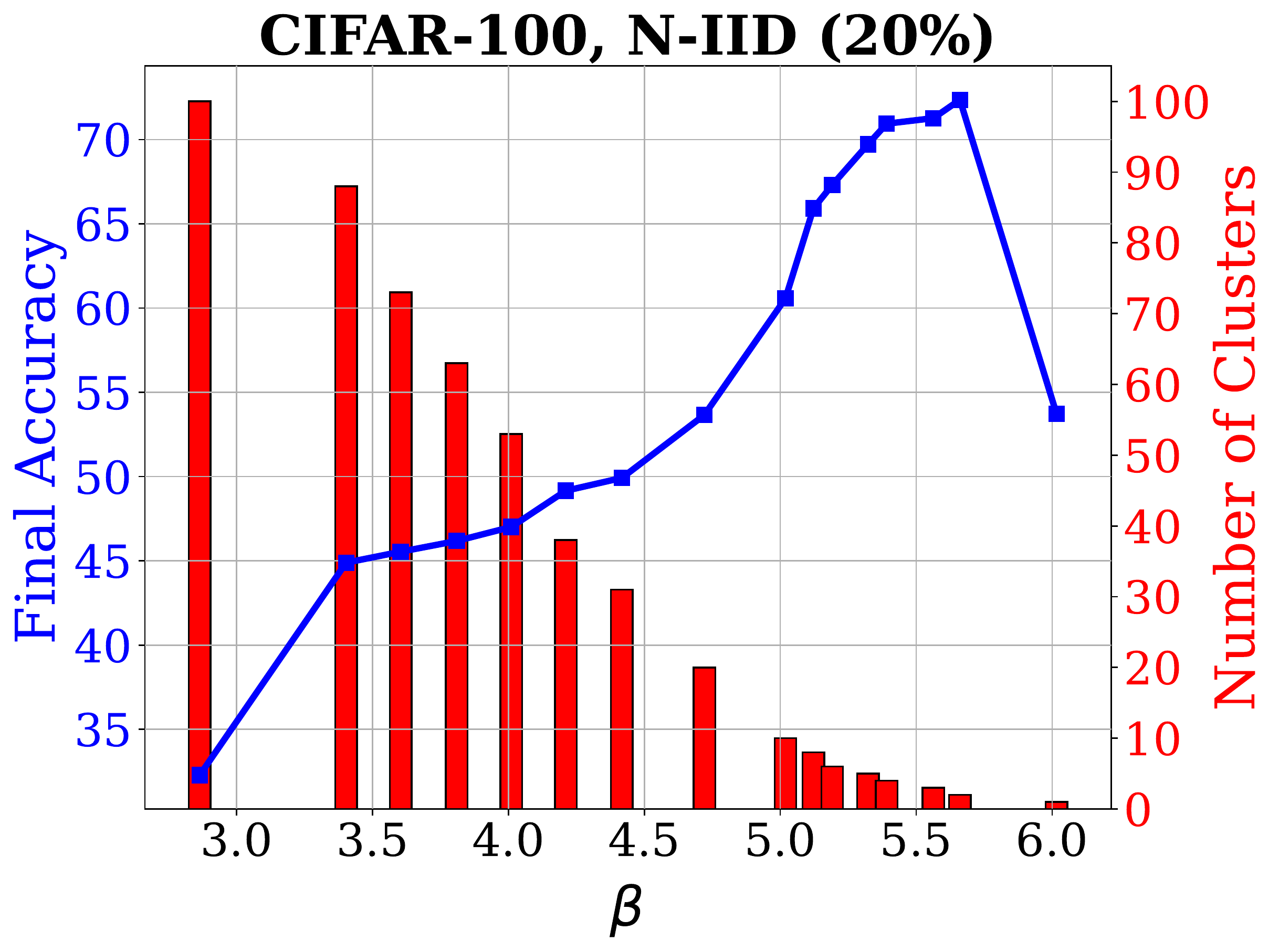}\quad
  \includegraphics[width=.18\pdfpagewidth]{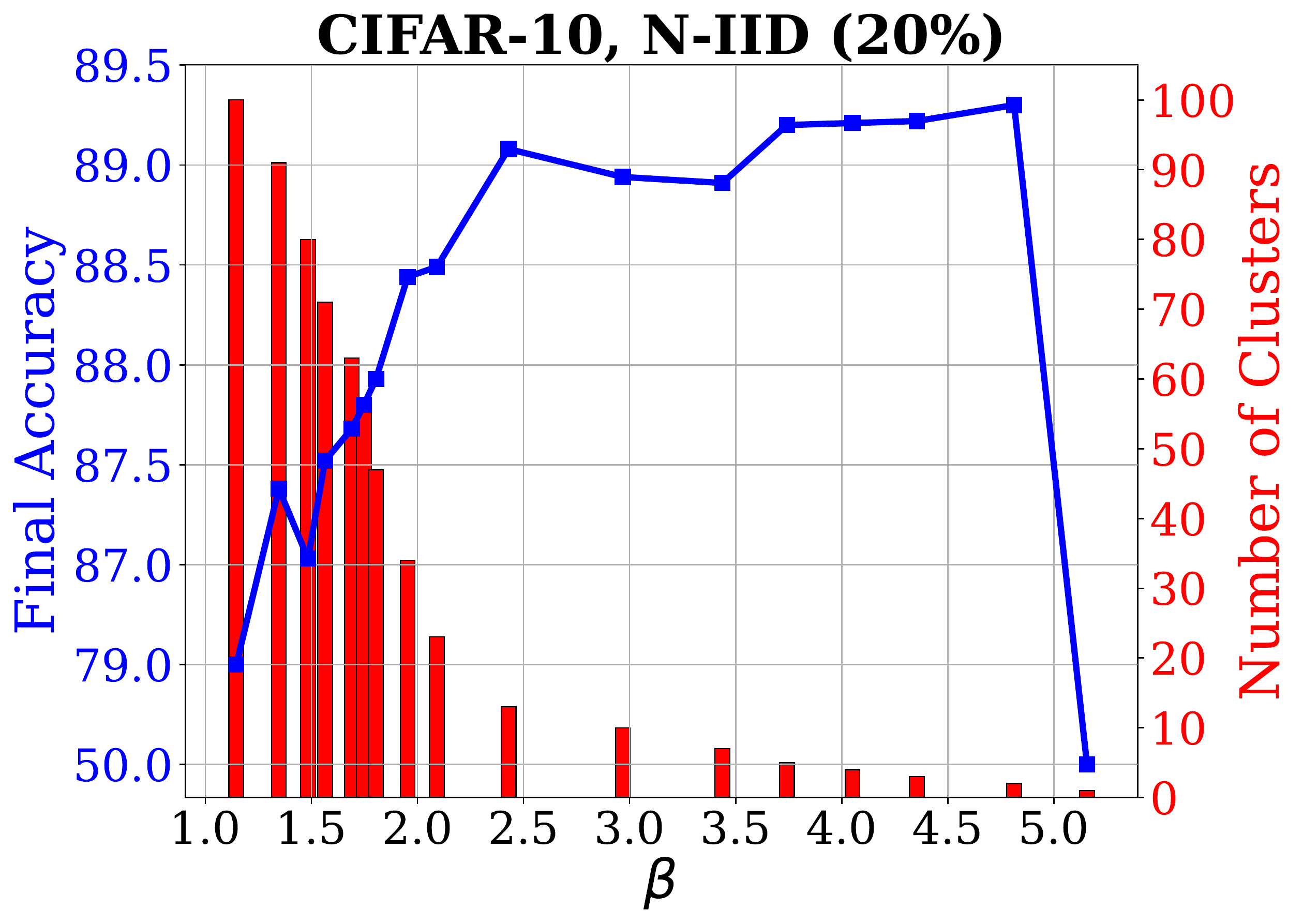}\quad
    \includegraphics[width=.18\pdfpagewidth]{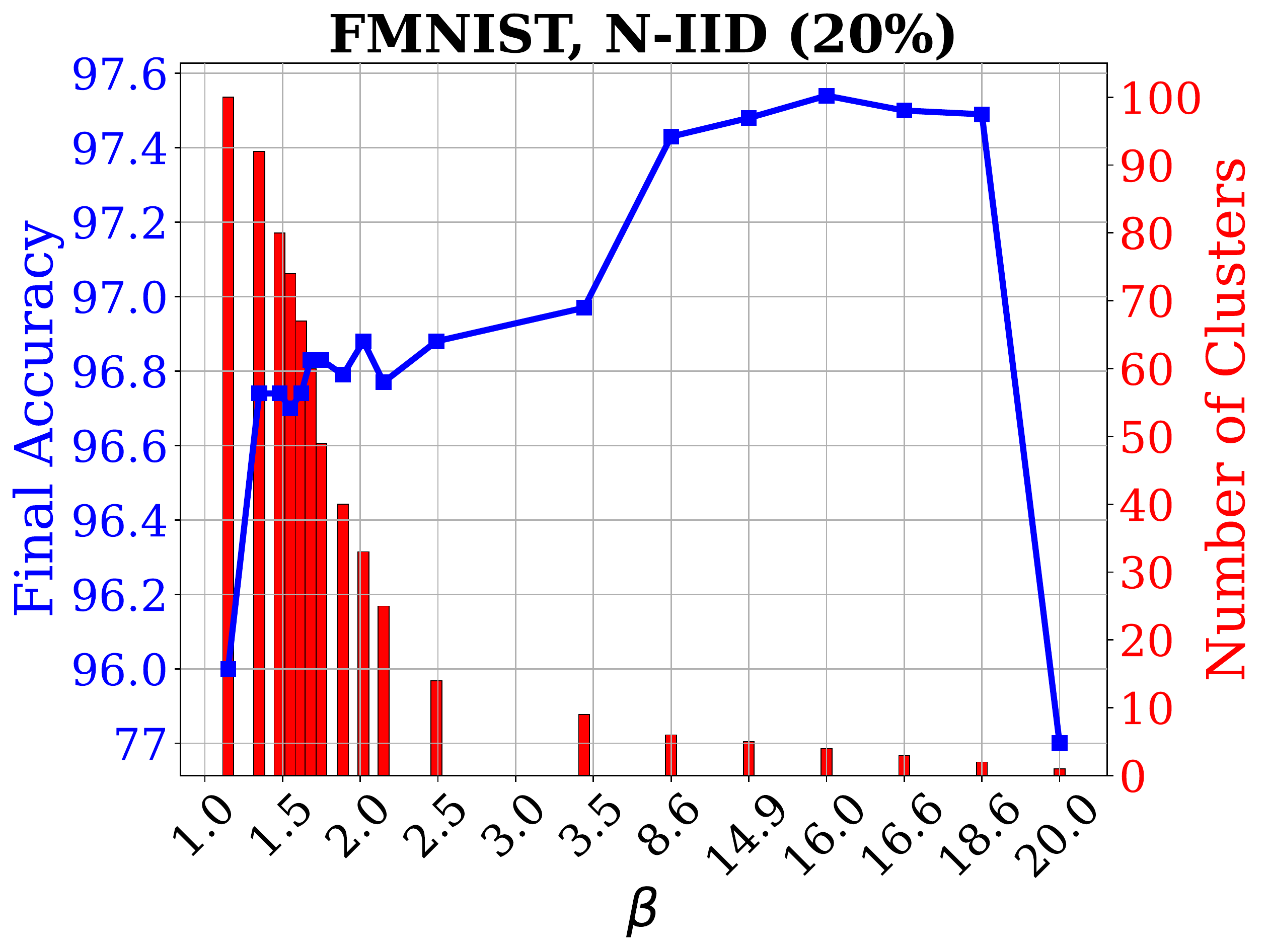}\quad
    \includegraphics[width=.18\pdfpagewidth]{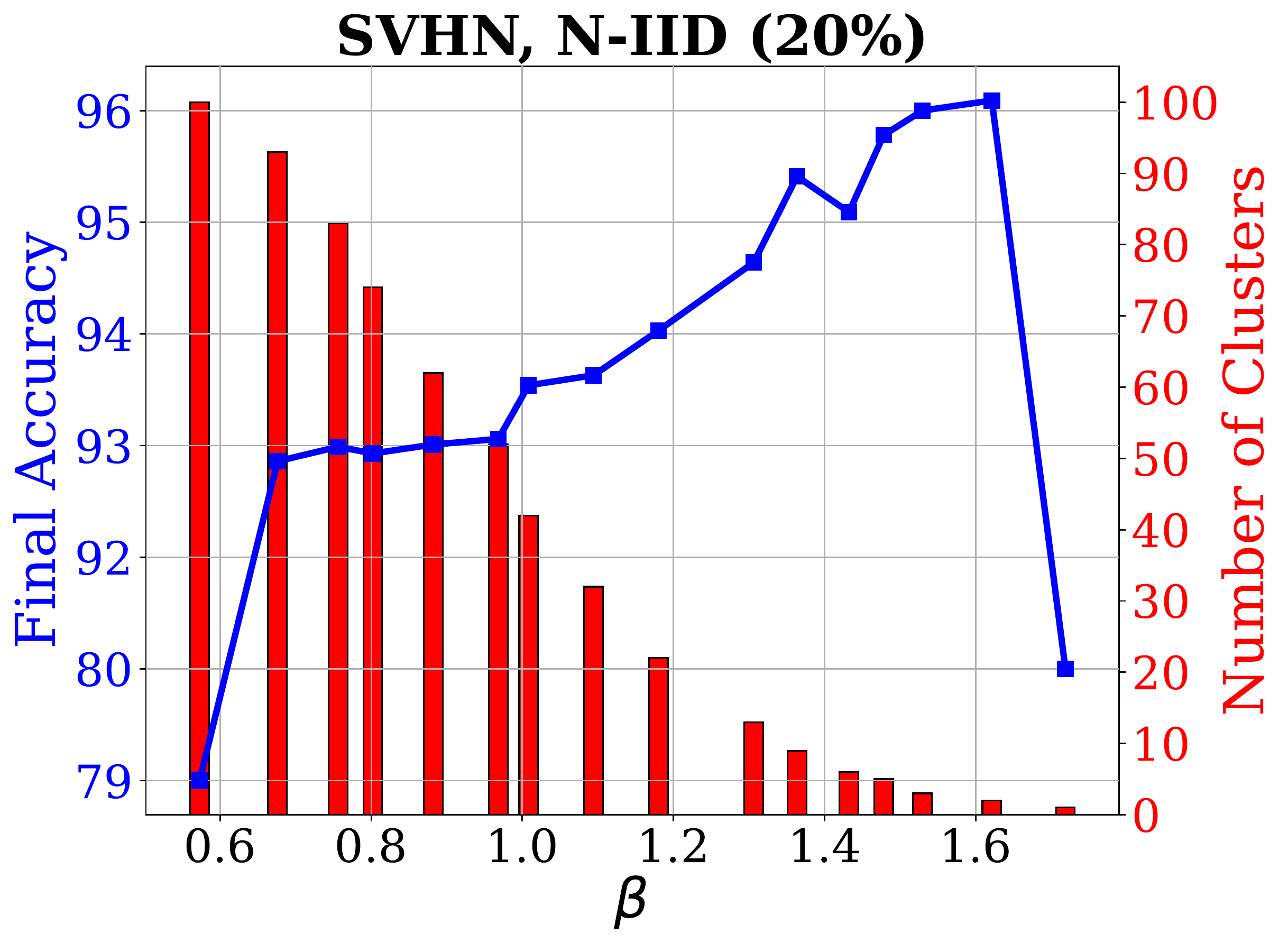}
    \vspace{-4mm}
    \caption{\footnotesize{Test accuracy performance of PACFL versus the clustering threshold $\beta$ (when the proximity matrix obtained  as in Eq.~\ref{adj1}), and the number of fitting clusters for Non-IID label skew ($20\%$) on CIFAR-10/100, FMNIST, and SVHN datasets. Each point in the plots are obtained by 200 communication rounds with local epoch of 10, local batch size of 10 and SGD local optimizer.}}
    \label{fig:beta20}
  \end{minipage}\\[1em]
  \vspace{-10mm}
\end{figure*}



\textbf{Datasets and Models}. We use image classification task and 4 popular datasets, i.e., FMNIST~\cite{xiao2017fashion}, SVHN~\cite{netzer2011reading}, CIFAR-10~\cite{krizhevsky2009learning}, CIFAR-100~\cite{krizhevsky2009learning}, to evaluate our method. For all experiments, we consider LeNet-5~\cite{lecun1989backpropagation} architecture for FMNIST, SVHN, and CIFAR-10 datasets and ResNet-9~\cite{he-2016-PAMI} architecture for CIFAR-100 dataset. 



\textbf{Baselines and Implementation}. To assess the performance of the proposed method against the SOTA, we compare PACFL against the following set of baselines.
For baselines that train a single global model across all clients, we compare with 
FedAvg~\cite{mcmahan2017communication}, ~FedProx~\cite{fedprox-smith-2020} ~FedNova~\cite{FedNova-2020}, and ~SCAFFOLD~\cite{scaffold-2020}.
For SOTA personalized FL methods, the baselines include~LG-FedAvg~\cite{LG2020}, Per-FedAvg~\cite{fallah2020personalized}, Clustered-FL (CFL)~\cite{sattler-clustered-fl-2021}, and IFCA~\cite{Ghosh-federated-2020}.
In all experiments, we assume 100 clients are available and 10$\%$ of them are sampled randomly at each round. Unless stated otherwise, throughout the experiments, the number of communication rounds is 200 and each client performs 10 locals epochs with batch size of 10 and local optimizer is SGD. We let $p$ in $\mathbf{U}_p$ be 3-5. 
Please refer to the supplementary material for more details about the experimental setup. 


  \vspace{-3mm}
\subsection{Overall Performance} \label{overall-performance}
 \vspace{-1mm}
We compare PACFL with all the mentioned SOTA baselines for two different widely used Non-IID settings, i.e. Non-IID label skew, and Non-IID Dirichlet label skew~\cite{Qinbin-Experimental-Study-2021}. We present the results of Non-IID label skew in the main paper and that of the Non-IID Dirichlet label skew in the supplementary material.  We report the mean and standard deviation for the average of final local test accuracy across all clients over 3 runs.

\noindent \textbf{Non-IID Label Skew.} In this setting, we first randomly assign $\varrho\%$ of the total available labels of a dataset to each client and then randomly distribute the samples of each label amongst clients own those labels as in~\cite{li2021federated}. In our experiments we use Non-IID label skew 20$\%$, and 30$\%$, i.e. $\varrho=\{20, 30\}\%$ respectively. Table~\ref{tab:niid2} shows the results for Non-IID label skew 20$\%$. We report the results of Non-IID label skew 30$\%$ in Tabel~\ref{tab:niid3} of the supplementary material. As can be seen, global FL baselines, i.e. FedAvg, FedProx, FedNova, and SCAFFOLD perform very poorly. That's due to weight divergence and model drift issues under heterogeneous setting~\cite{zhao2018-non-iid}. We can observe from Table~\ref{tab:niid2} that PACFL consistently outperforms all SOTA on all datasets. In particular, focusing on CIFAR-100, PACFL outperforms all SOTA methods (by $+19\%, +18\%, +19\%, +18\%$ for FedAvg, FedProx, FedNova, SCAFFOLD)  as well as all the personalized competitors (by $+27\%, +13\%, +1.5\%, +33\%$ for LG, PerFedAvg, IFCA, CFL ). We tuned the hyperparameters in each baseline to obtain the best results. IFCA achieved the best performance with 2 clusters which is consistent with the results in~\cite{Ghosh-federated-2020}.

\footnotesize{
\begin{table}
\footnotesize

\begin{center}
\footnotesize

 \caption{\footnotesize{Test accuracy comparison across different datasets for Non-IID label skew $(20\%)$. For each baseline, the average of final local test accuracy over all clients is reported. We run each baseline 3 times for 200 communication rounds with local epoch of 10.}}
\vspace{-4mm}
    \color{black}
    \label{tab:niid2}
    \centering
\scalebox{0.78}{
\begin{tabular} {p{1.1cm}p{1.71cm}p{1.71cm}p{1.71cm}p{1.71cm}}
            \toprule
            Algorithm & FMNIST & CIFAR-10 & CIFAR-100 & SVHN\\
            \midrule
            SOLO & $95.92 \pm 0.57$ & $79.22 \pm 1.67$  & $32.28 \pm 0.23$  & $79.72 \pm  1.37$\\
            FedAvg   & $77.3 \pm 4.9$ & $49.8 \pm 3.3$ & $53.73 \pm 0.50$ & $80.2 \pm 0.8$\\
            FedProx  & $74.9\pm 2.6$ & ${50.7 \pm 1.7}$ & $54.35 \pm 0.84$ & $79.3 \pm 0.9$\\
            FedNova   & $70.4 \pm 5.1$ & $46.5 \pm 3.5$ & $53.61 \pm 0.42$ & $75.4 \pm 4.8$\\
            Scafold    & $42.8 \pm 28.7$ & $49.1 \pm 1.7$ & $54.15 \pm 0.42$ & $62.7 \pm 11.6$\\
            LG & $96.80 \pm 0.51$ & $86.31 \pm 0.82$ & $45.98 \pm 0.34$ & $92.61 \pm 0.45$\\
            PerFedAvg  & $95.95 \pm 1.15$ & $85.46 \pm 0.56$ & $60.19 \pm 0.15$ & $93.32 \pm 2.05$\\
            IFCA &  $97.15 \pm 0.01$ & $87.99 \pm 0.15$ & $71.84 \pm 0.23$ & $95.42 \pm 0.06$\\
            CFL  & $77.93 \pm 2.19$ & $51.11 \pm 1.01$ & $40.29 \pm 2.23$ & $73.62 \pm 1.76$\\
            \rowcolor{LightCyan} \textbf{PACFL}  & $\bf{97.54 \pm 0.08}$ & $\bf{89.30 \pm 0.41}$ & $\bf{73.10 \pm 0.21}$ & $\bf{95.77 \pm 0.18}$\\
            \midrule
            
        \end{tabular}
       } 

\end{center}
  \vspace{-9mm}
  
 \end{table}
}

\normalsize


\normalsize

\normalsize



\begin{figure*}

    \centering
    \begin{subfigure}[b]{0.2\textwidth}
         \centering
         \includegraphics[width=.13\pdfpagewidth, height=.12\pdfpagewidth]{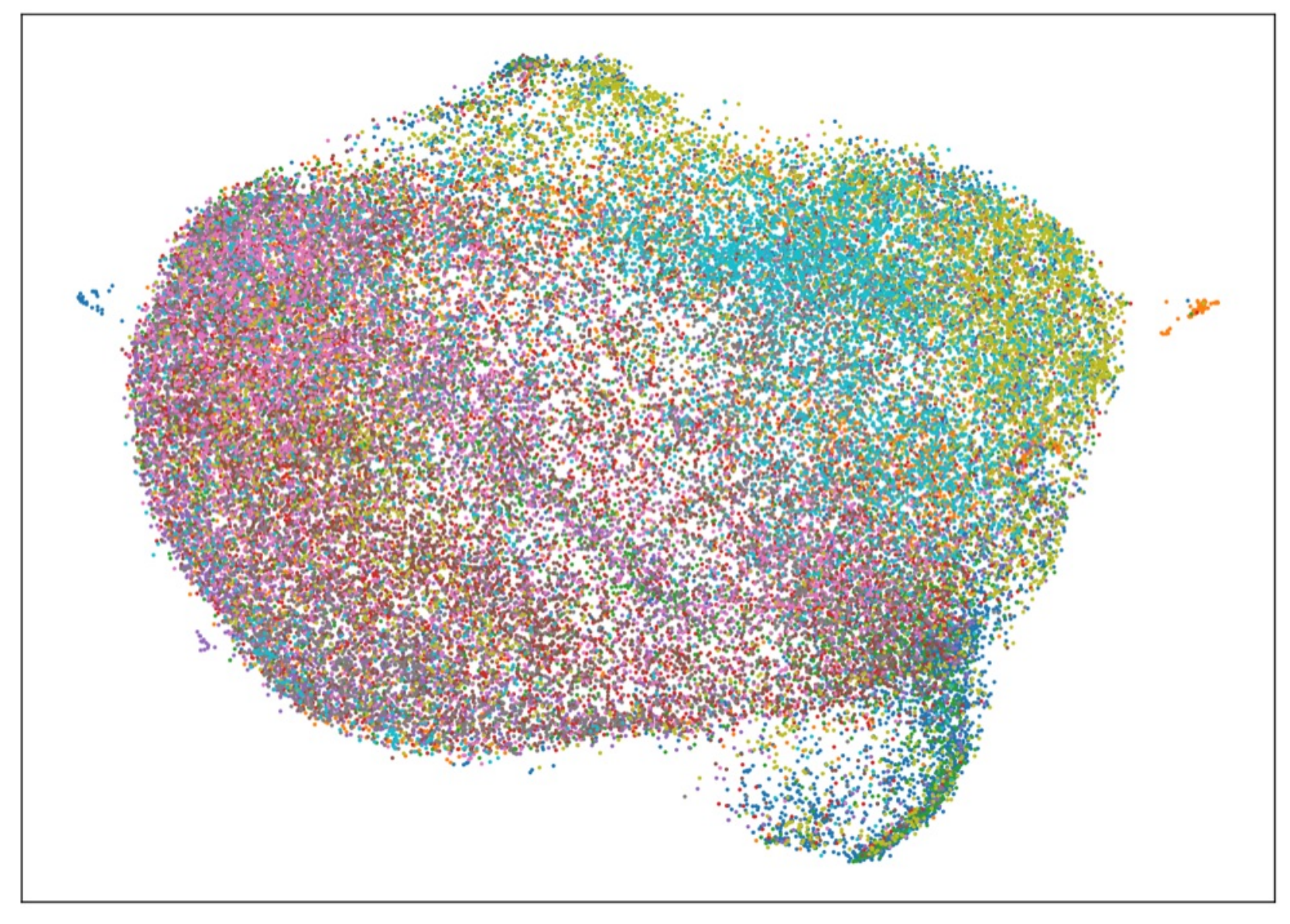}
         \caption{}
         \label{fig:umap-adj1-a}
     \end{subfigure}
     \hfill
     \begin{subfigure}[b]{0.2\textwidth}
         \centering
         \includegraphics[width=.13\pdfpagewidth, height=.12\pdfpagewidth]{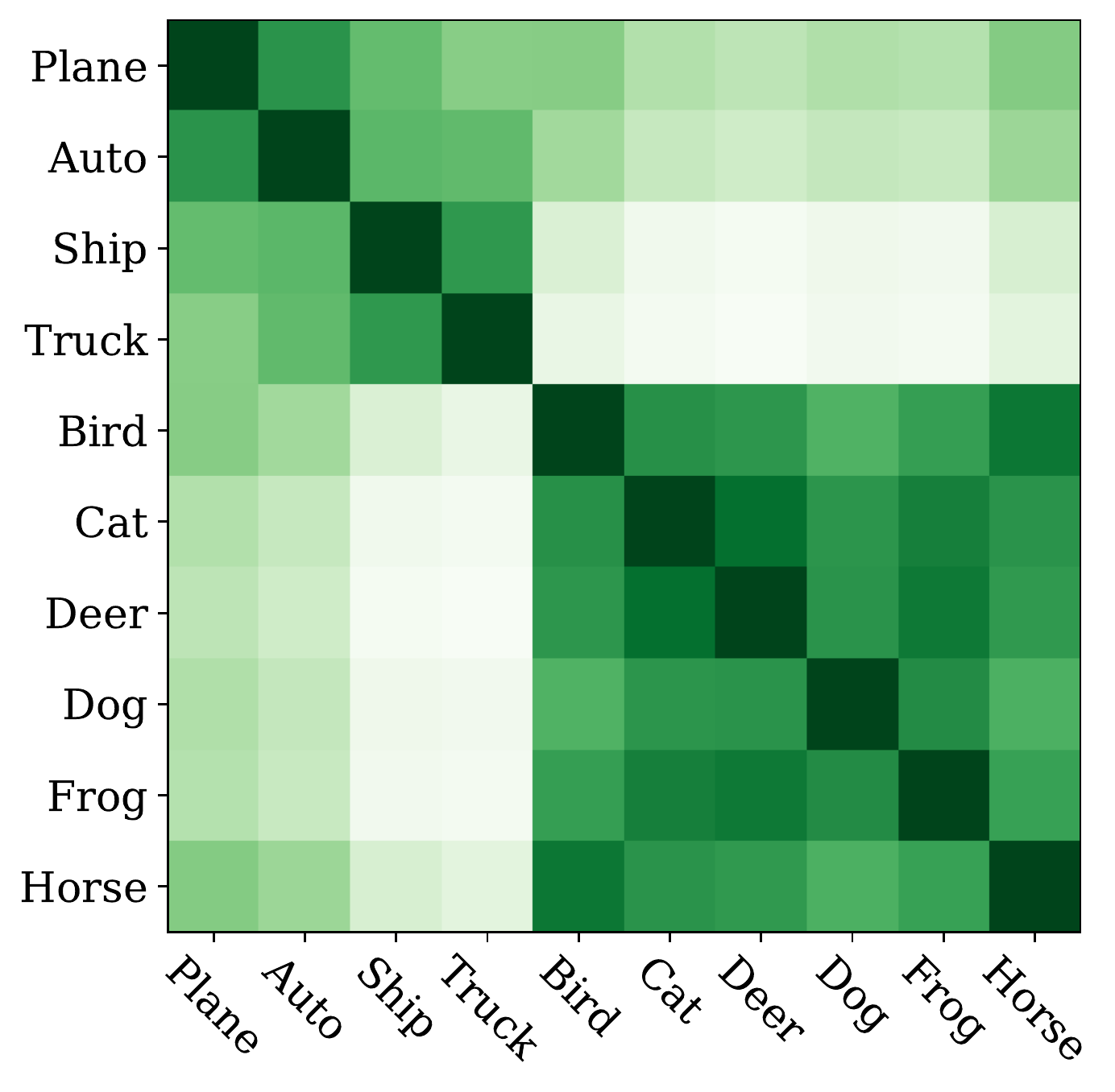}
         \caption{}
         \label{fig:umap-adj1-b}
     \end{subfigure}
     \hfill
     \begin{subfigure}[b]{0.2\textwidth}
         \centering
         \includegraphics[width=.13\pdfpagewidth, height=.12\pdfpagewidth]{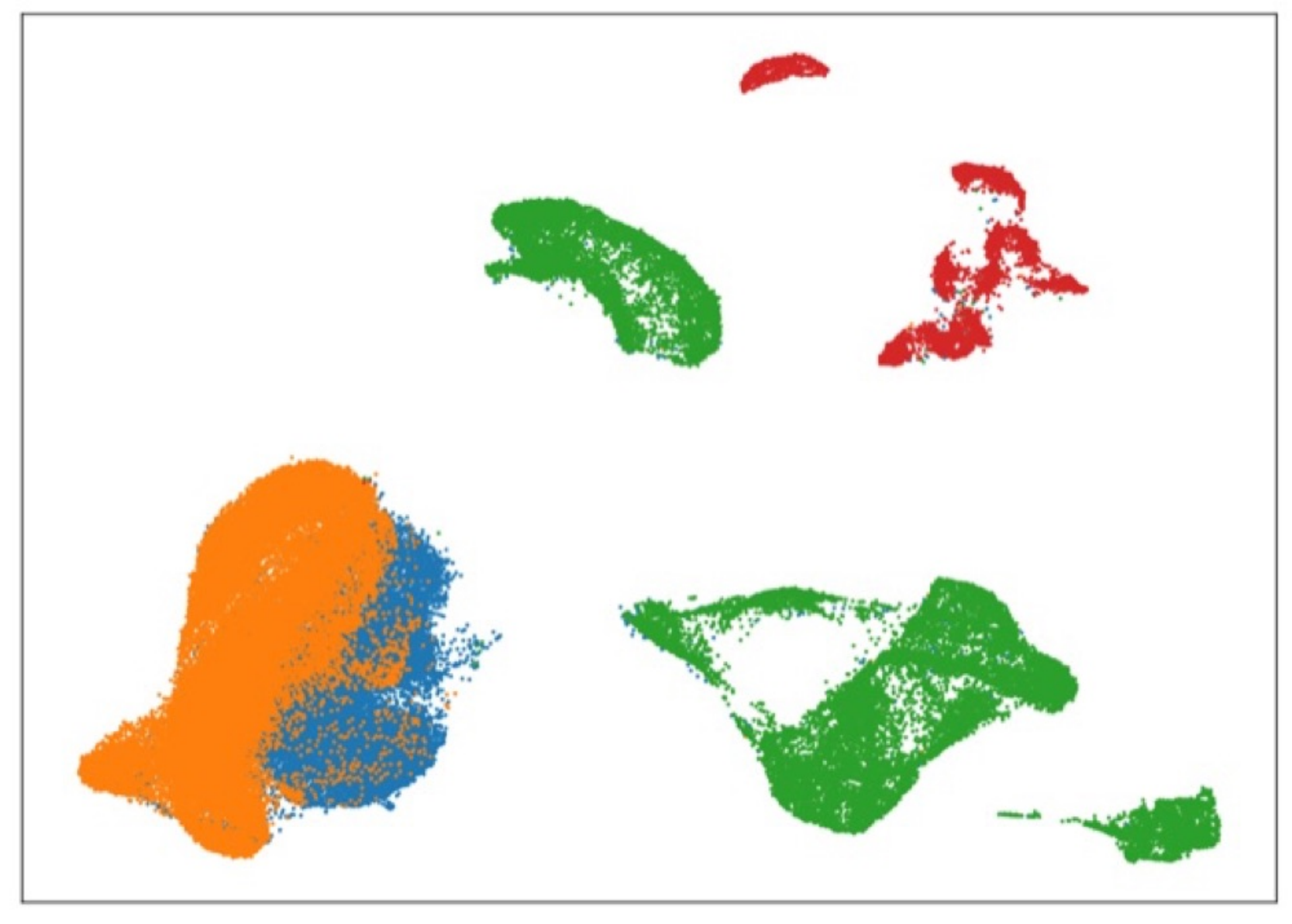}
         \caption{}
         \label{fig:umap-adj1-c}
     \end{subfigure}
     \hfill
      \begin{subfigure}[b]{0.2\textwidth}
         \centering
         \includegraphics[width=.13\pdfpagewidth, height=.12\pdfpagewidth]{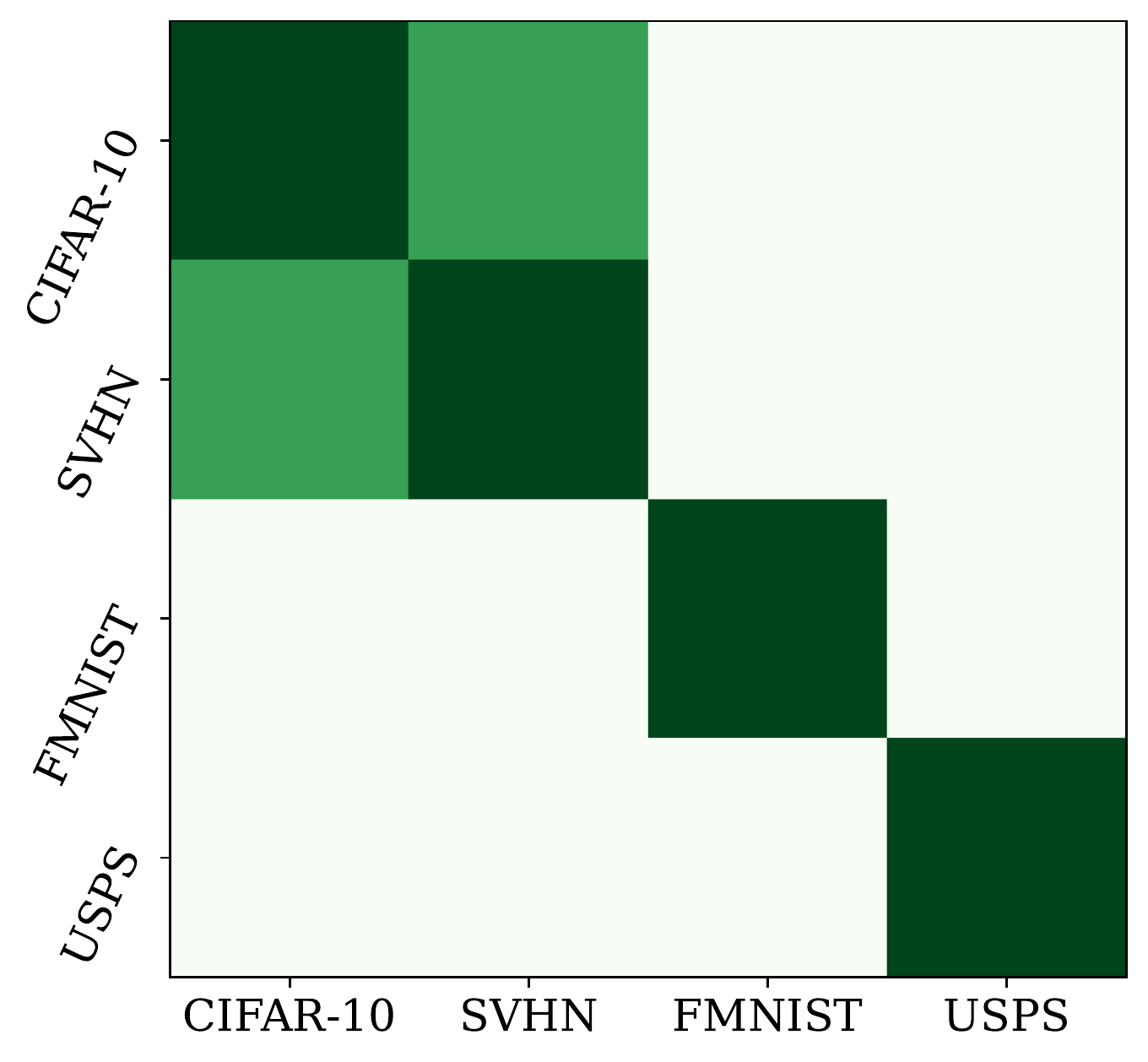}
         \caption{}
         \label{fig:umap-adj1-d}
     \end{subfigure}
  \vspace{-4mm}
    \caption{\footnotesize{The main message of this visualization is to understand the cluster structure of different datasets as well as the distribution similarity of different heterogeneous tasks/datasets. (a) depicts the UMAP visualization of CIFAR-10 classes. As can be seen, CIFAR-10 naturally has two super clusters, namely animals (cat, dog, bird, deer, horse, frog)  and vehicles (car, plane, ship, truck), which are shown in the purple and green regions, respectively. This means that within each super cluster, the distance between the distribution of the classes is small. While the distance between the distributions of the two super clusters are quite huge. Since the union of clients data is CIFAR-10, two cluster is enough to handle the Non-IIDness across clients. This is the reason that the best accuracy performance on CIFAR-10 is obtained when the number of clusters is 2. (b) We obtained the proximity matrix $\mathbf{A}$ as in Eq.~\ref{adj1} and sketched it. The entries of $\mathbf{A}$ are the smallest principle angle between all pairs of classes of CIFAR-10. This concurs with (a) showing the cluster structure of CIFAR-10. (c) The data of MIX-4 is naturally clustered into four clusters.  The structure of the 4 clusters is also accurately suggested by PACFL for this task. (d) We did the same thing as in (b) for MIX-4 as well and sketched the matrix.}}
    \label{fig:umap-adj1}
    \vspace{-6mm}
\end{figure*}


\vspace{-3mm}
\subsection{Globalization and Personalization Trade-off}
\vspace{-1mm}
\label{Global-personal-trade-off}

To cope with the statistical heterogeneity, previous works incorporated a proximal term in local optimization or modified the model aggregation scheme at the server side to take the advantage of a certain level of personalization~\cite{fedprox-smith-2020, Vahidian-federated-2021, mahdavi2020-APFL}. Though effective, they lack the flexibility to trade off between \emph{personalization} and \emph{globalization}. Our proposed PACFL approach can naturally provide this globalization and personalization trade-off. Fig.~\ref{fig:beta20} visualizes the accuracy performance behavior of PACFL versus different values of $\beta$ which is the $L_2$ (Euclidean) distance between two clusters when the proximity matrix obtained  as in Eq.~\ref{adj1}, or Eq.~\ref{adj2}. In other words, $\beta$ is a threshold controlling the number of clusters as well as  the similarity of the data distribution of clients within a cluster under Non-IID label skew. The blue curve and the red bars demonstrate the accuracy, and the number of clusters respectively for each $\beta$. Varying $\beta$ in a range which depends upon the dataset, PACFL can sweep from training a fully global model (with only 1 cluster) to training fully personalized models for each client.

As is evident from Fig.~\ref{fig:beta20}, the behaviour of PACFL on each dataset is similar. In particular, increasing $\beta$, decreases the number of clusters (by grouping more number of clients within each cluster and sharing more training) which realizes more globalization. When $\beta$ is big enough, PACFL will group all clients into 1 cluster and the scenario reduces to the FedAvg baseline (pure globalization). On the contrary, decreasing $\beta$, increases the number of clusters, which leads to more personalization. When $\beta$ is small enough, individual clusters would be formed for each client and the scenario degenerates to the SOLO baseline (pure personalization). As demonstrated, on all datasets, all clients benefit from some level of globalization. 
This is the reason why decreasing the number of clusters can improve the accuracy performance in comparison to SOLO. 
In general, \emph{finding the optimal trade-off between globalization and personalization depends on the level of heterogeneity of tasks, the intra-class distance of the dataset, as well as the data partitioning across the clients}.
This is precisely what PACFL is designed to do, to find this optimal trade-off
before initiating the federation via the proximity matrix at the server. IFCA lacks this trade-off capability as it must define a fixed number of clusters $(C > 1)$ or with $C = 1$ it would degenerate to FedAvg.

\vspace{-3mm}
\subsection{Mixture of 4 Datasets} \label{main-mix4}
\vspace{-1mm}

Existing studies have been evaluated on simple partitioning strategies, i.e., Non-IID label skew $(20\%)$ and $(30\%)$~\cite{li2021federated}. While these partitioning strategies synthetically simulate Non-IID data distributions in FL by partitioning a dataset into multiple smaller Non-IID subsets, they cannot design  real and challenging Non-IID data distributions. According to the prior sections, due to the small intra-class distance (similarity between distribution of the classes) in the used benchmark datasets, all baselines benefited highly from globalization. This is the reason that PACFL and IFCA could achieve a high performance with only 2 clusters. 

In order to better assess the potential of the SOTA baselines under a real-world and challenging Non-IID task where the local data of clients have strong statistical heterogeneity, we design the following experiment naming it as MIX-4. We assume that each client owns data samples from one of the four datasets, i.e., USPS~\cite{hull1994database}, CIFAR-10, SVHN, and FMNIST. In particular, we distribute CIFAR-10, SVHN, FMNIST, USPS among 31, 25, 27, 14 clients respectively (100 total clients) where each client receives 500 samples from all classes of only one of these dataset. This is a hard Non-IID task. We compare our approach with the SOTA baselines in the classification of these four different datasets, and we present the average of the clients' final local test accuracy in~Table~\ref{tab:mix4}. 
As can be seen, IFCA is unable to effectively handle this difficult scenario with tremendous data heterogeneity with just two clusters, as suggested in~\cite{Ghosh-federated-2020} as the best fitting number of clusters. 
IFCA (2) with 2 clusters performs almost as poorly as the global baselines while PACFL can find the optimal number of clusters in this task (four clusters) and outperforms all SOTA by a large margin. The results of IFCA (4) with 4 clusters is $76.79 \pm 0.43$. As observed, PACFL surpasses all the global competitors (by $+14\%, +15\%, +16\%, +8\%$ for FedAvg, FedProx, FedNova, SCAFFOLD)  as well as all the personalized competitors (by $+19\%, +35\%, +7\%, +16\%$ for LG, PerFedAvg, IFCA, CFL, respectively). Further, the visualization in Fig.~\ref{fig:umap-adj1-c} and \ref{fig:umap-adj1-d} also show how PACFL determines the optimal number of clusters on MIX-4.

\begin{table}
\begin{center}
\caption{\footnotesize{The benefits of PACFL are particularly pronounced when the tasks are extremely Non-IID. This table evaluates different FL approaches in the challenging scenario of MIX-4 in terms of the top-1 test accuracy performance. 
While all competing approaches have substantial difficulties in handling this scenario with tremendous data heterogeneity,
the results clearly show that PACFL is very robust even under such difficult data heterogeneity scenarios. 
}}
\vspace{-4mm}
\label{tab:mix4}

\footnotesize{
 \scalebox{0.8}{
        \begin{tabular}{p{2cm}p{2cm}||p{2cm}p{2cm}}
            \toprule
            Algorithm & MIX-4 &  Algorithm & MIX-4 \\
            \midrule
SOLO & $55.08 \pm 0.29$  &LG& $58.49 \pm 0.46$ \\
FedAvg & $63.68 \pm 1.64$ & PerFedAvg & $42.60 \pm 0.60$ \\
FedProx & $61.86 \pm 3.73$ & IFCA $(2)$ & $70.32 \pm 3.57$\\
FedNova & $60.92 \pm 3.60$ & CFL & $61.18 \pm 2.63$\\
Scaffold & $69.26 \pm 0.84$ & $ \textbf{PACFL}$ & $\bf{77.83 \pm 0.33}$\\

            \midrule
\end{tabular}
     }
    }
\end{center}
\vspace{-7mm}
\end{table}

\normalsize


\begin{figure*}[htbp]
  \begin{minipage}{\textwidth}
    \centering
   \includegraphics[width=.15\pdfpagewidth]{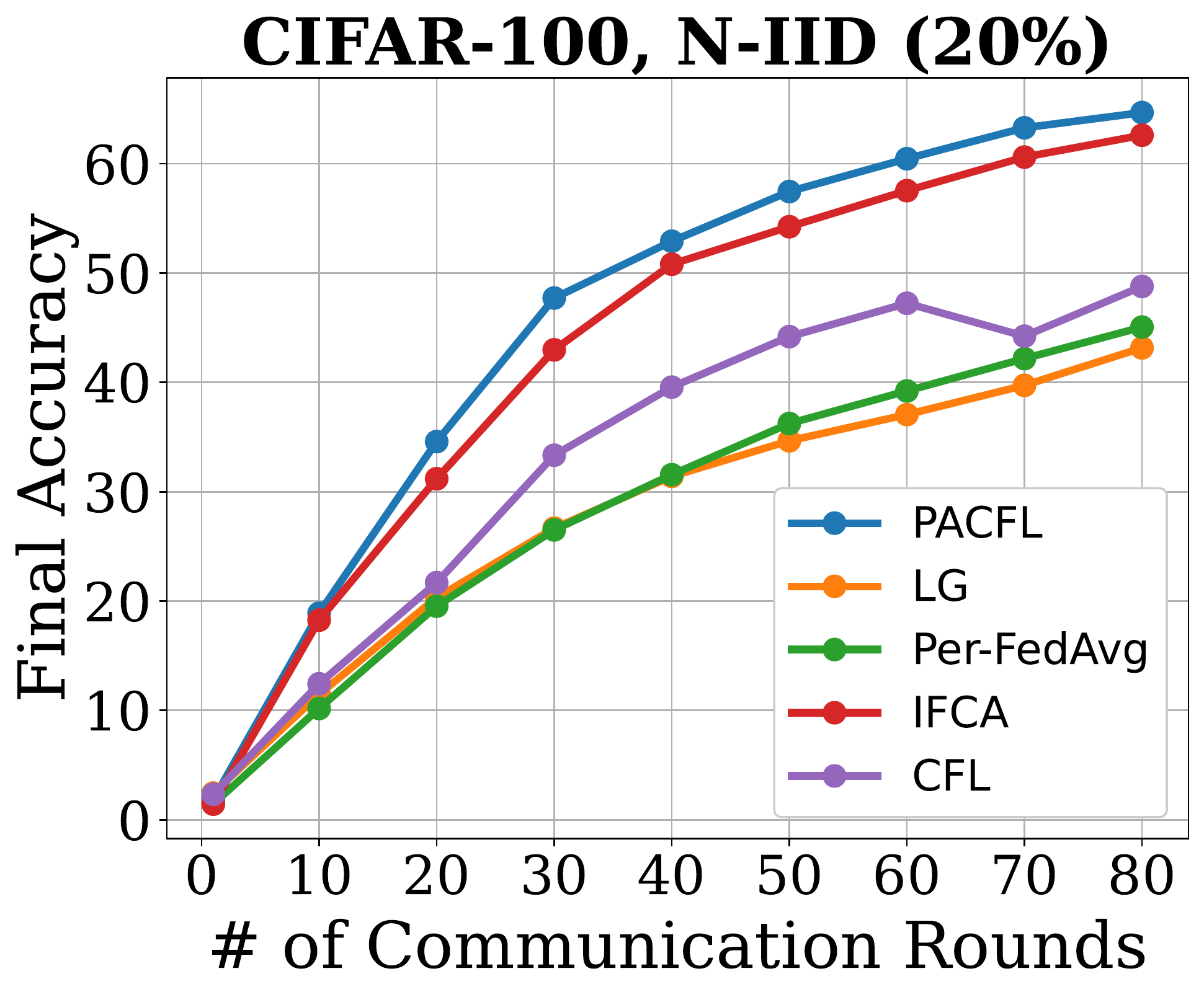}\quad
   \includegraphics[width=.15\pdfpagewidth]{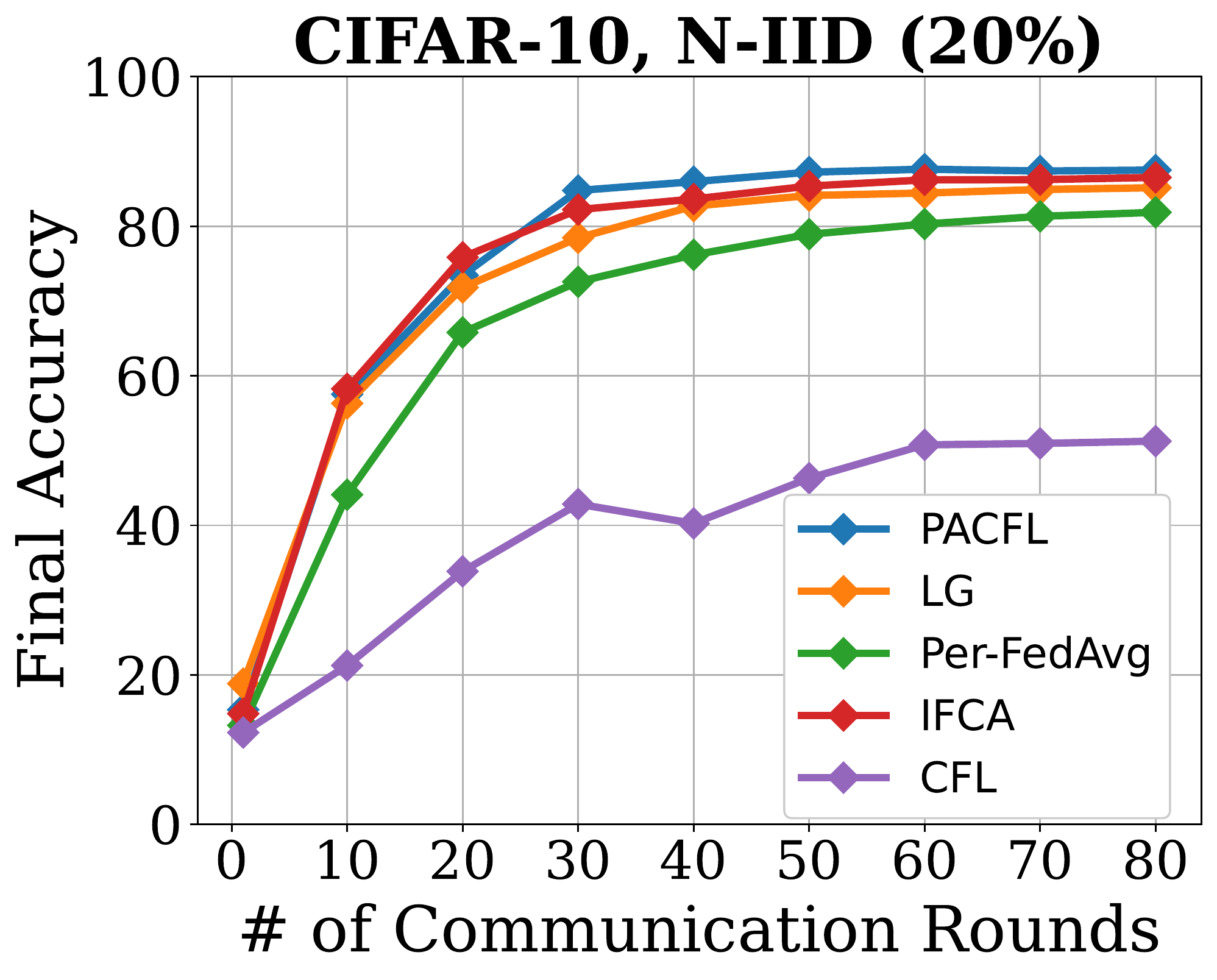}\quad
    \includegraphics[width=.15\pdfpagewidth]{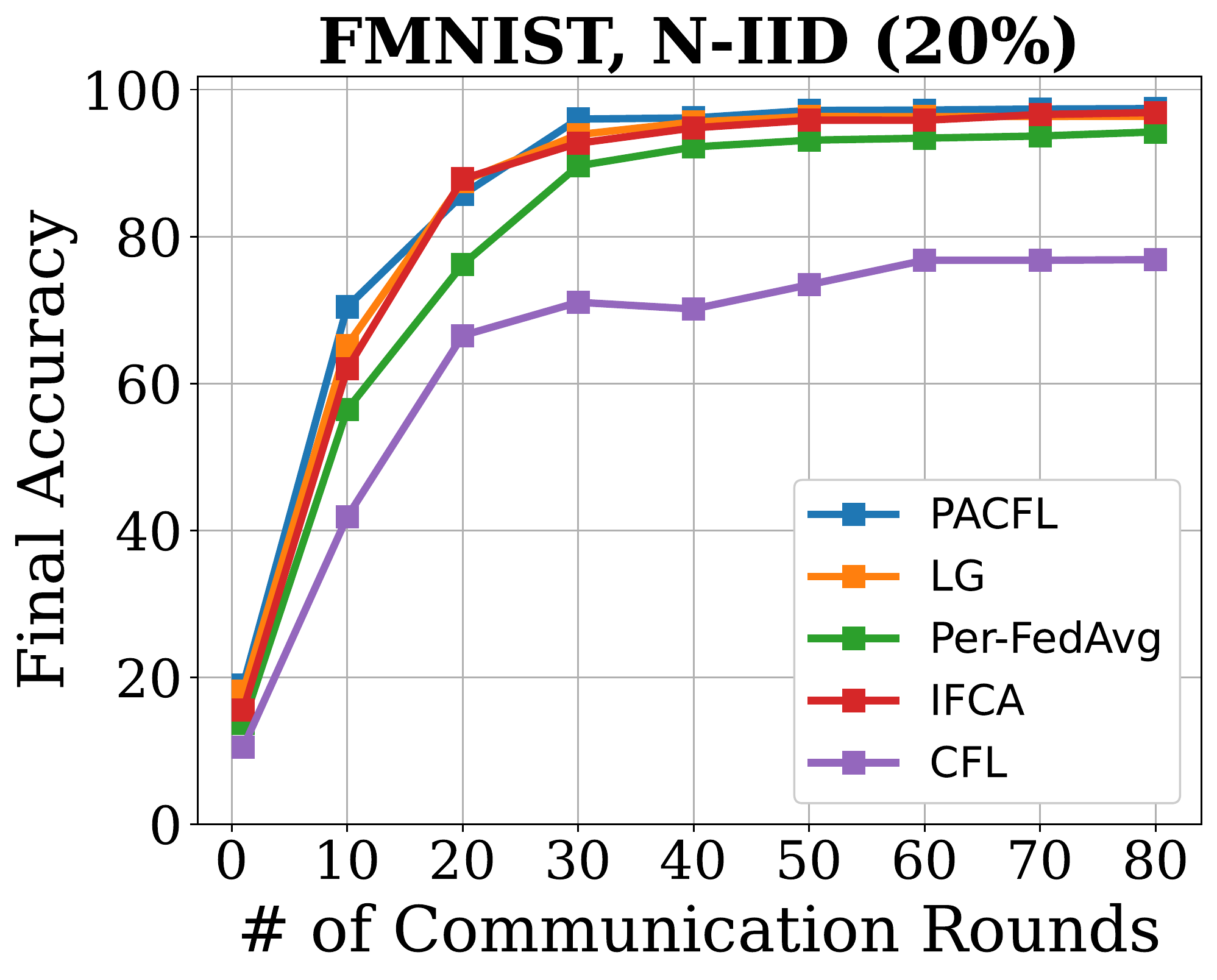}\quad
    \includegraphics[width=.15\pdfpagewidth]{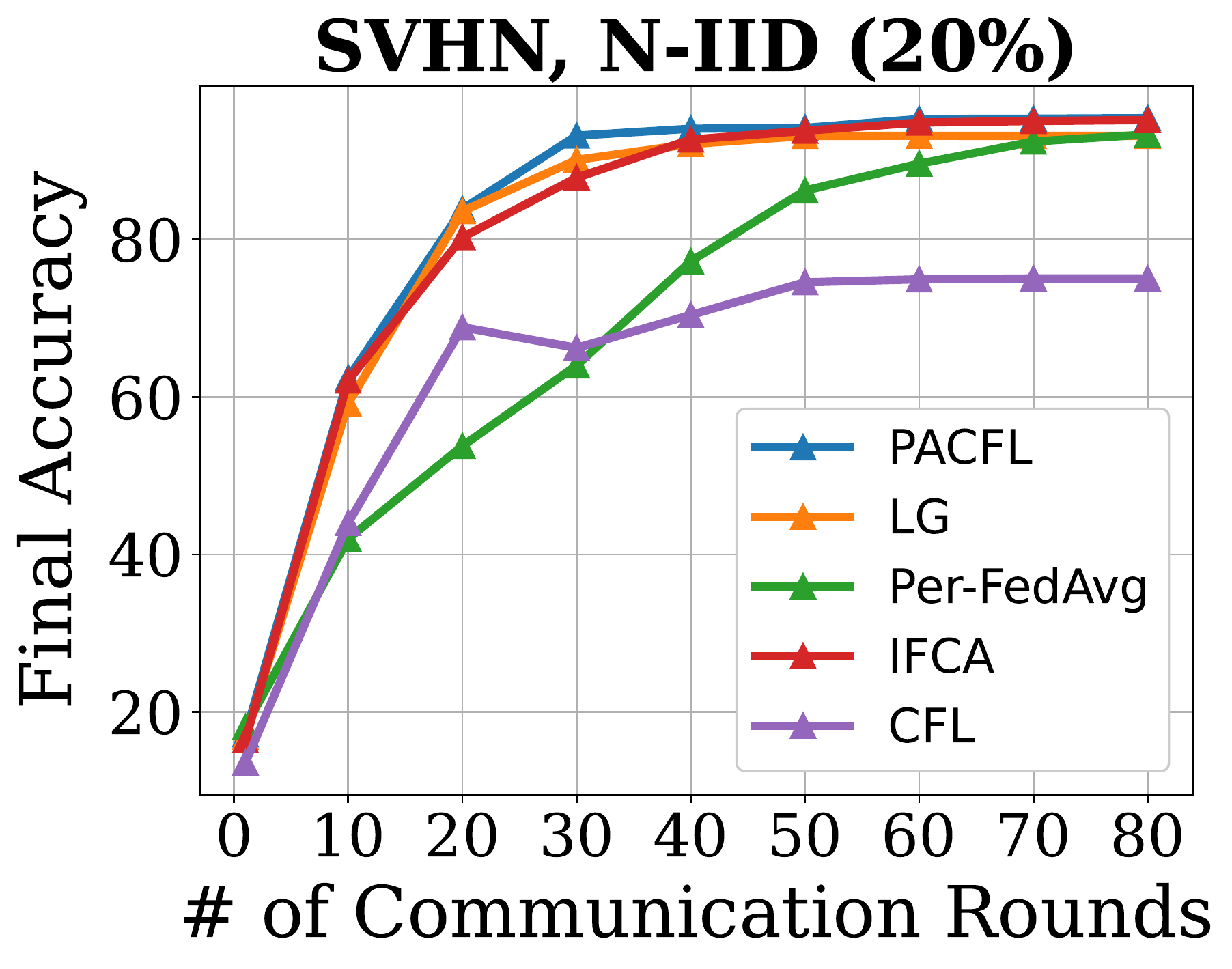}
    \vspace{-4mm}
    \caption{Test accuracy versus number of communication rounds for Non-IID ($20\%$). PACFL converges fast to the desired accuracy and consistently outperforms strong competitors.}
    \label{fig:comm-cost-niid2}
  \end{minipage}\\[1em]
  \vspace{-11mm}
\end{figure*}

 \vspace{-3mm}
\subsection{How Many Clusters Are Needed?}
\label{How Many Clusters Are Needed}
 \vspace{-1mm}


As we emphasized in prior sections, one of the significant contributions of PACFL is that the server can easily determine the best fitting number of clusters just by analyzing the proximity matrix without running the whole federation. For instance, for IID scenarios, we expect the best fitting number of clusters to be one. The reason behind is that under IID setting, since all clients have similar distributions, they can share in the training of the averaged global model to benefit all. On the other hand, 
in the case of MIX-4,
we expect the best fitting number of clusters to be four.
More generally, we expect the best fitting number of clusters to be dependent on the similarities/dissimilarities in distributions among the clients.
We empirically show in Fig.~\ref{fig:beta20} that the best accuracy results on CIFAR-100, CIFAR-10, SVHN, and FMNIST for Non-IID label skew (20$\%$) are obtained when the number of clusters are 2, 2, 2, and 4, respectively. 

The UMAP~\cite{mcinnes2018umap} visualization in Fig.~\ref{fig:umap-adj1-a}
also confirms that two clusters is the best case for training the local models on partitions of CIFAR-10 dataset. Broadly speaking, CIFAR-10 has 2 big classes, i.e., class of animals (cat, dog, deer, frog, horse, bird) and class of vehicles (airplane, automobile, ship and truck). Fig.~\ref{fig:umap-adj1-b} also depicts the proximity matrix of CIFAR-10 dataset, whose entries are the principal angle between the subspace of every pairs of 10 classes (labels). This further confirms that our proposed method perfectly captures the level of heterogeneity, thereby finding the best fitting number of clusters in a privacy preserving manner. In particular, our experiments demonstrate that the clients that have the sub-classes of these two big classes have common features and can improve the performance of other clients that own sub-classes of the same big class if they are assigned to the same cluster. A similar observation can be seen for other datasets.

In PACFL, the server only requires to receive the \emph{signature} of the clients data in one-shot and thereby initiating the federation with the best fitting number of clusters. This translates to several orders of magnitude in communication savings for PACFL. However, as mentioned in~\cite{Ghosh-federated-2020}, IFCA treats the number of clusters as a hyperparameter which is optimized after running the whole federation with different number of clusters which increases the communication cost by several orders of magnitude.

\normalsize
 \vspace{-1mm}
\subsection{Generalization to Newcomers }
\label{Generalization-to-Newcomers}
\vspace{-1mm}
PACFL provides an elegant approach to handle \emph{newcomers} arriving after the federation procedure, to learn their personalized model. In general, for all other baselines it is not clear how they can be extended to handle clients unseen at training (federation). We will show in Algorithm~\ref{alg:newcomers} (in the supplementary material) how PACFL can simply be generalized to handle clients arriving after the end of federation, to learn their personalized model. The unseen client will send the signature of its data to the server and the server determines which cluster it belongs to. The server then sends the corresponding model to the newcomer and the newcomer fine tunes the received model. To evaluate the quality of newcomers' personalized models, we design an experiment under Non-IID label skew (20$\%$) where only 80 out of 100 clients are participating in a federation with 50 rounds. Then, the remaining 20 clients join the network at the end of federation and receive the model from the server and personalize it for only 5 epochs. The average final local test accuracy of the unseen clients is reported in Table~\ref{tab:unseen-main}. Focusing on CIFAR-100, as observed, some of the personalized baselines including LG and PerFedAvg perform as poor as global baselines and SOLO. PACFL consistently outperforms other baselines by a large margin.

\normalsize

\vspace{-3mm}
\begin{table}[htbp]
\caption{Average local test accuracy across unseen clients' on different datasets for Non-IID label skew $(20\%)$.}
 \vspace{-4mm}
\centering
\scalebox{0.75}{
\begin{tabular}{lllll}
\toprule
Algorithm & FMNIST & CIFAR-10 & CIFAR-100 & SVHN \\
\midrule
SOLO & $95.13 \pm 0.42$ & $82.30 \pm 1.00$ & $27.26 \pm 0.98$ & $91.5 \pm 0.64$ \\
FedAvg & $77.61 \pm 3.78$ & $31.01 \pm 1.83$ & $32.19 \pm 0.32$ & $71.78 \pm 3.43$ \\
FedProx & $74.30 \pm 4.70$ & $27.56 \pm 3.24$ & $32.41 \pm 1.17$ & $74.30 \pm 4.70$ \\
FedNova & $74.66 \pm 2.81$ & $31.48 \pm 1.49$ & $33.18 \pm 0.80$ & $73.04 \pm 3.65$ \\
Scafold & $73.97 \pm 1.68$ & $37.22 \pm 1.34$ & $23.90 \pm 2.61$ & $64.96 \pm 4.74$ \\
LG & $94.58 \pm 0.33$ & $77.98 \pm 1.61$ & $10.63 \pm 0.21$ & $89.48 \pm 0.65$ \\
PerFedAvg & $89.88 \pm 0.38$ & $73.79 \pm 0.51$ & $30.09 \pm 0.35$ & $67.48 \pm 2.88$ \\
IFCA & $96.29 \pm 0.04$ & $84.98 \pm 0.41$ & $55.66 \pm 0.20$ & $94.83 \pm 0.14$ \\
\rowcolor{LightCyan} \textbf{PACFL} & $\bf{96.36 \pm 0.20}$ & $\bf{87.14 \pm 0.15}$ & $\bf{59.16 \pm 0.42}$ & $\bf{95.25 \pm 0.08}$ \\
\bottomrule
\end{tabular}
}
\label{tab:unseen-main}
\vspace{-6.5mm}
\end{table}

\vspace{-1mm}
\subsection{Communication Cost}\label{comm-cost-section}
\vspace{-1mm}
\textbf{Learning with Limited Communication} In this section we consider circumstances that frequently arise in practice, where a limited amount of communication round is permissible for federation under a heterogeneous setup. To this end, we compare the performance of the proposed method with the rest of SOTA. Herein, we consider a limited communication rounds budget of $80$ for all personalized baselines and present the average of final local test accuracy over all clients versus number of communication rounds for Non-IID label skew (20$\%$) in Fig.~\ref{fig:comm-cost-niid2}. Our proposed method requires only 30 communication rounds to converge in CIFAR-10, SVHN, and FMNIST datasets. CFL yields the worst performance on all benchmarks across all datasets, except for CIFAR-100. Per-Fedavg seems to benefit more from higher communication rounds. IFCA, is the closest line to ours for CIFAR-10, SVHN and FMNIST, however, PACFL consistently outperforms IFCA. This can be explaiend by the fact that IFCA randomly initializes cluster models that are inherently noisy, and many rounds of federation is required until the formation of clusters is stabilized. Further, IFCA is sensitive to initialization and a good initialization of cluster model parameters is key for convergence~\cite{FLT2021}. This issue can be further pronounced by the results presented in Table~\ref{tab:comm-20} which demonstrate the number of communication round required to achieve a designated target accuracy. In this table, ``{\footnotesize $--$}'' means that the baseline is unable to reach the specified target accuracy. As can be seen, PACFL beats IFCA and all SOTA methods. \vspace{-2mm}

\begin{table}[htbp]
    \color{black}
    \caption{\footnotesize{Comparing different FL approaches for Non-IID ($20\%$) in terms of the required number of communication rounds to reach target top-1 average local test accuracy.}}
    \vspace{-3mm}
    \label{tab:comm-20}
    \centering
    {\footnotesize
    \scalebox{0.8}{

        \begin{tabular}{p{10mm}p{15mm}p{15mm}p{15mm}p{8mm}}
            \toprule
            Algorithm & FMNIST & CIFAR-10 & CIFAR-100 & SVHN\\
            \midrule
            Target  & $75$ & $80$ & $50$ & $75$ \\
            \midrule
            FedAvg   & $200$ & $--$ & $130$ & $150$\\
            FedProx   & $200$ & $--$ & $115$ & $200$\\
            FedNova   & $--$ & $--$ & $120$ & $150$\\
            Scafold    & $--$ & $--$ & $--$ & $--$\\
            LG  & $13$ & $33$ & $--$ & $16$\\
            PerFedAvg  & $19$ & $60$ & $110$ & $39$\\
            IFCA  & $14$ & $25$ & $40$ & $17$\\
            CFL  & $47$ & $--$ & $--$ & $--$\\
\rowcolor{LightCyan} \textbf{PACFL}  & $\bf{12}$ & $\bf{24}$ & $\bf{37}$ & $\bf{15}$\\
            \bottomrule
        \end{tabular}
        }
    }
    
    \vspace{-7mm}
\end{table}

\normalsize

\section{Conclusion}
  \vspace{-1mm}
In this paper, we proposed a new framework for clustered FL that directly aims to efficiently identify distribution similarities among clients by analyzing the principal angles between the client data subspaces spanned by their principal vectors. This approach provides a simple, but yet effective clustered FL framework that addresses a broad range of data heterogeneity issues beyond simpler forms of Non-IIDness like label skews.

\clearpage

\newpage
\setcounter{section}{0}
{\Large \textbf{Supplementary Document}}

The supplementary material is organized as follows. Section~\ref{Preliminaries-supmat} provides additional preliminaries; Section~\ref{Convergence-supmat}
presents theoretical results regarding the convergence of PACFL; Section~\ref{toy example} shows the consistency of our approach with several well-known distribution similarity metrics; Section~\ref{ Background and Related Work}
presents additional background and related work; Section~\ref{Extra Experimental Results}
provides additional experiments to evaluate the performance of the proposed approach; finally, Section~\ref{imp}
contains implementation details.




\section{Preliminaries} \label{Preliminaries-supmat}
This section complements section~\ref{Preliminaries} of the paper.


\noindent \textbf{Truncated Singular Value Decomposition (SVD).} Truncated SVD of a real $m\times n$ matrix M is a factorization of the form $\mathbf { \tilde M}=\mathbf {U}_p \mathbf{\Sigma}_p \mathbf{V}^{T}_p$ where $ \mathbf{U}_p=\left[ \mathbf {\mathbf{u}_1, \mathbf{u}_2,..., \mathbf{u}_p} \right]$ is an $m\times p$  orthonormal matrix, $\mathbf{\Sigma_p}$ is a $p\times p$ rectangular diagonal matrix with non-negative real numbers on the main diagonal, and $\mathbf {V}_p=\left[  \mathbf{v}_1, \mathbf{v}_2,..., \mathbf{v}_p \right]$ is a $p\times n$ orthonormal matrix, where $\mathbf {u}_i \in \mathbf {U}_p$ and $\mathbf {v}_i \in \mathbf {V}_p$ are the left and right singular vectors, respectively.

\vspace{\medskipamount}

\noindent \textbf{Hierarchical Clustering.}
When constituting disjoint clusters where the number of clusters is not known a priori, hierarchical clustering (HC)~\cite{day1984efficient} is of interest. Agglomerative HC is one of the popular clustering techniques in machine learning that takes a proximity (adjacency) matrix as input and groups similar objects into clusters. Initially each data point is considered as a separate cluster. At each iteration, it repeatedly do the following two steps: (1) identify the two clusters that are the closest, and (2) merge the two most similar clusters. This iterative process continues until one cluster or $C$ clusters are formed. In order to decide which clusters should be merged, a linkage criterion such as single, complete, average, etc is defined~\cite{day1984efficient}. For instance, in ``single linkage", the pairwise $L_2$ (Euclidean) distance between two clusters is defined as the smallest distance between two points in each cluster. In this paper, $\beta$ represents the distance between two clusters which we call it clustering threshold.
 \vspace{-1mm}
\section{Convergence Analysis}
\label{converg-anal}
\vspace{-1mm}

We take the perspective of concentrating on one cluster and considering the set of clients assigned to it as performing Local SGD. Since clients are selected at random and data is Non-IID, we adapt the convergence framework presented in~\cite{haddadpour2019convergence} and consider how the presence of potentially misidentified clients (meaning a client's data truly belongs in cluster $j$ but is assigned to cluster $k$) affect the convergence guarantees. We show that the rate does not change, and only the order constant is increased by a quantity proportional to the maximum gradient norm of the loss associated with the misidentified clients.

Formally, we define the SGD training as seeking to solve the optimization problem,
\[
\min\limits_{\theta} \sum_{k\in C} F_k(\theta)
\]

We now place PACFL along the lines of theoretical analysis associated with~\cite{haddadpour2019convergence}.
This paper analyzed the convergence guarantees of distributed local SGD algorithms with Non-IID data, however with some measure
of the similarity or dissimilarity of the data. Using this framework, we consider the convergence of the global parameter set $\theta^t_{g,z}$ for some cluster $z$ on the sum of losses corresponding to the data on the clients associated with that cluster. The measure introduced in~\cite{haddadpour2019convergence} to indicate data distributions that are not identical but similar is called \emph{Weighted Gradient Diversity}, defined as, \vspace{-2mm}
\begin{equation}\label{eq:gradsim}
\Lambda(\theta,C) := \frac{\sum_{k\in C} \left\|\nabla F_k(\theta)\right\|^2}{|C|\left\|\sum_{k\in C} \nabla F_k(\theta)\right\|^2}
\end{equation}
where we have defined their clients' model parameters to be equal, i.e., $q_k=1/|C|$ for all $k$ and now included the dependence on $|C|$.

At each round, a set of clients is randomly chosen and a sequence of SGD steps are
taken, and then they are averaged across all of the clients, and then the local procedures start again. This procedure is
exactly the same as the classic local SGD, as analyzed in~\cite{haddadpour2019convergence}, with two primary modifications:
\begin{enumerate}
    \item There are several concurrent clusters performing independent local SGD, with the clusters
defined by the similarity measure. Since generally speaking the structure of the behavior is the same across clusters, an analysis on one is representative. 
\item Since clustering is a statistical and noisy procedure, i.e., there is no guarantee that the clusters are \emph{correct}, we must perform the analysis based on the understanding that there exists a decomposition of the clients included in that cluster
as correctly or incorrectly identified. In particular correct identification implies that the bound in Eq.~\ref{eq:gradsim} holds, and incorrect identification implies no guaranteed similarity in the gradients. We note that this is a worst-case assumption -- even if the data is distinct enough to be clustered, we would expect some inter-cluster similarity, even if it's less than intra-cluster similarity.
\end{enumerate}   

In the analysis we shall drop the subscript $z$ identifying the cluster (i.e., $\theta^t_{g,z}=\theta^t_g$ considering one as representative. With this we introduce notation as follows: 
\begin{enumerate}
    \item $c$ is the total number of clients available associated to the cluster.  
    
    \item There are $E$ local SGD steps (the ClientUpdate step in PACFL) that come with the client update. 
   
    \item Each SGD step for every client is taken as a minibatch of size $B$, allowing for variance control.  
   
    \item We now consider $t$ the counter of the total number of SGD steps rather than ``major" averaging iterations (rounds), as in the description of PACFL.  
   
    \item There are $K$ randomly chosen clients belonging to the cluster at each iteration.
\end{enumerate}
This places the algorithm squarely in the framework of local SGD for Non-IID data, when we restrict consideration to the performance associated with just one of the clusters. The one modification we intend to study to this framework is the presence of misidentified clients; due to statistical noise there are some members of the cluster that do not have similar data, and thus do not exhibit gradient similarity. Formally, we make the following assumption regarding correct and incorrect identification,
\begin{assumption}\label{as:cluster}
Let us consider that cluster that clients $\bar{C}=\{1,...,\bar{c}\}$ have been correctly identified, i.e., $\Lambda(\theta, C)\le \lambda$. Furthermore,
assume that there exists $U_g$ such that
\begin{equation}\label{eq:gradbound}
\|\nabla F_k(\theta) \|\le U_g
\end{equation}
for all $k\in C\setminus \bar{C}=\{\bar{c}+1,...,c\}$ and $\theta\in\{\{\theta^t_g\},\{\theta^t_k\}\}$.
\end{assumption}
Thus, we consider the case of $c-\bar{c}$ incorrectly identified clients whose gradients are effectively garbage to the optimization process for the cluster of interest. We will investigate the degree to which they slow down or bias the optimization procedure.
Of course, a priori we cannot expect the gradient to be bounded, especially for objectives with a quadratic growth (e.g. PL inequality) property, however, in practice one can see the dependence of the convergence result below on $U_g$ as indicating the effect misidentified clusters present for the convergence rate. Naturally, if a misidentified client's gradients are effectively garbage for a cluster, the effect on the convergence will be dependent on how large these gradients can get.

A significant contribution of PACFL is the use of the \emph{data} rather than the loss value as a mechanism of cluster identification. This distinction from works such as~\cite{Ghosh-federated-2020} permits analysis to be performed for non-convex objectives, uniquely in the recent cluster literature. Let $F(\theta)=\sum_{k\in C} F_k(\theta)$. Recall that $\theta^t_g$ are the parameter values averaged across the
clients associated to the cluster and let $g_g^t$ be the averaged stochastic gradients at iteration $t$ while $g_j^t$ are the individual clients' stochastic gradients. We define each SGD iteration as:
\begin{equation}\label{eq:sgd}
    \theta^{t+1}_j = \theta^t_j-\eta_t g_j^t
\end{equation}
where $g^t_j$ is a stochastic gradient. We present some problem and algorithmic assumptions below.

\vspace{-2mm}
\begin{assumption}\label{as:problem}
\begin{enumerate}
    \item Each $F_k$ is $L$-smooth, i.e., $\|\nabla F_k(\theta)-\nabla F_k(\theta')\|\le L \|\theta-\theta'\|$
    \item $F(\theta)$ is lower bounded by $f^*$
    \item $F(\theta)$ satisfies the Polyak-\L ojasiewicz (PL) inequality with constant $\mu$, i.e., $\frac{1}{2}\|\nabla F(\theta)\|^2\ge \mu (F(\theta)-F(\theta^*))$
    \item The stochastic gradient estimates are unbiased and have a variance bound according to,  \vspace{-1mm}
    \[
    \mathbb{E}\left[\|g_j^t-\nabla F_j(\theta_j^t)\|^2\right]\le C_1 \|\nabla F_j(\theta_j^t)\|^2+\frac{\sigma^2}{B}
    \]
    where $\sigma$ is a positive constant reflecting the individual sample variance.
\end{enumerate}
\end{assumption}

We now present the main convergence results. Their proofs, which involve minor modifications of~\cite{haddadpour2019convergence}, are in Section~\ref{Convergence-supmat} of the supplementary material.
For the first result, let $\kappa=L/\mu$, the ratio of the largest to smallest (throughout the objective landscape) eigenvalue of the Hessian of the objective, be the condition number of the problem.
\begin{theorem}
\label{th:convergence}
There exist constants $a$ and $B$ depending on the problem and $\lambda$ such that for sufficiently large $E$, number of local stochastic gradient updates, and $K$, number of sampled nodes from each cluster, it holds that, after
$T$ total iterations,  \vspace{-1mm}
\begin{equation}\label{eq:convergence}
\begin{array}{l}
    \mathbb{E}\left[F(\theta^T_g)-F^*\right]
    \le \frac{a^3}{(T+a)^3}\mathbb{E}\left[F(\theta^0_g)-F^*\right] 
    + \\ 
    \frac{4\kappa \sigma^2 T(T+2a)}{\mu K B(T+a)^3}+\frac{256 \kappa^2 \sigma^2 T(E-1)}{\mu KB(T+a)^3}
    \end{array}
\end{equation}
This matches the standard rate for objectives with quadratic growth for local SGD, i.e.,  \vspace{-1mm}
\[
\mathbb{E}\left[F(\theta^T_g)-F^*\right] = O\left(\frac{1}{KT}\right)
\]
\end{theorem}
The proof is in subsection~\ref{proof-th1} of the supplementary material.
For the general non-convex objectives we have the following result:
\vspace{-1mm}
\begin{theorem}\label{th:convnonconvex}
Let a constant stepsize $\eta$ satisfy,
\[
-\frac{\eta}{2}+\frac{\lambda(K+1)L^2 \eta^3[2 C_1+E(E+1)]}{2K}+\frac{\lambda L^2\eta^2}{2}\left(\frac{C_1}{K}+1\right)\le 0
\]
It holds that,  \vspace{-1mm}
\[
\begin{array}{l}
\frac{1}{T}\sum\limits_{t=0}^{T-1}\mathbb{E}\|\nabla F(\theta^t_g)\|^2 \le \frac{2(F(\theta^t_g)-f^*)}{\eta T} + \\
\hspace{3cm} \frac{L\eta}{2}\left(\frac{\sigma^2}{KB}+\left(\frac{c}{K} + 1\right)U_g^2\right) + \\
\hspace{3cm} \frac{\eta^2 L^2(K+1)((E+1)\sigma^2+(C_1+E(E+1))U^2_g)}{KB}
\end{array}
\]
Thus, with stepsize $\eta = \frac{1}{L}\sqrt{K}{T}$, the standard sublinear convergence rate on the average expected gradient holds, i.e.,  \vspace{-1mm}
\[
\frac{1}{T}\sum\limits_{t=0}^{T-1}\mathbb{E}\|\nabla F(\theta^t_g)\|^2 = O\left(\frac{1}{\sqrt{KT}}\right)
\]
\end{theorem}

In summary, as long as there is some bound on all of the gradients as evaluated at the parameter values of the misidentified cluster, there is neither a bias in the solution nor a slowdown of the rate of convergence, simply an increase in the constant. We can interpret the convergence as indicating the degree to which being proportional to the square of the largest gradient norm across the parameter values at which a stochastic gradient is evaluated for these clients during training.


\subsection{Proof of Theorems} \label{Convergence-supmat}
In this part we provide the proofs of the theoretical results discussed in the main paper.
\subsubsection{Proof of Theorem~\ref{th:convergence}.} \label{proof-th1}
The proof follows along the same lines as~\cite[Theorem 4.2]{haddadpour2019convergence}. we proceed with the same structure with the minor adjustments as needed.

We start with a descent Lemma, whose proof is unchanged.  
\begin{lemma}\cite[Lemma B.1]{haddadpour2019convergence}\label{lem:desclem}
\[
\begin{array}{l}
\mathbb{E}\left[\mathbb{E}_{\mathcal{P}_t}[F(\theta^{t+1}_g)-F(\theta^t_g)]\right]\le  -\eta_t \mathbb{E}\left[\mathbb{E}_{\mathcal{P}_t}
[\langle \nabla F(\theta^t_g),g_g^t\rangle]\right] \\
\hspace{4.1cm} + \frac{\eta_t^2 L}{2}\mathbb{E}\left[\mathbb{E}_{\mathcal{P}_t}[\|\theta^t_g\|^2]\right]
\end{array}
\]
\end{lemma}
The next result is a minor alteration of~\cite[Lemma B.2]{haddadpour2019convergence}.
\begin{lemma}\label{lem:gsqbound}
\[
\begin{array}{l}
\mathbb{E}\left[\mathbb{E}_{\mathcal{P}_t}[\|g_g^t\|^2]\right] \le \left(\frac{c}{K}+1\right)\frac{1}{c}\left[\sum\limits_{j=1}^c 
\|\nabla F_j(\theta_{j}^t)\|^2\right]+\frac{\sigma^2}{KB} 
\le \\ \frac{\lambda}{c^2}\left(\frac{c}{K}\hspace{-1mm}+\hspace{-1mm}1\right)\left\|\sum\limits_{j=1}^{\bar{c}} \nabla F_j(\theta^t_j)\right\|^2 \hspace{-1mm}+\hspace{-1mm}\left(\frac{c}{K}\hspace{-1mm}+\hspace{-1mm}1\right)\frac{1}{c}\left[\sum\limits_{j=\bar{c}+1}^c 
\|\nabla F_j(\theta_j^t)\|^2\right] \\ +\frac{\sigma^2}{KB}
\end{array}
\]
\end{lemma}
The next statement remains unchanged;
\begin{corollary}\cite[Corollary B.4]{haddadpour2019convergence}
\[
\begin{array}{l}
-\eta_t\mathbb{E}\left[\mathbb{E}[\langle \nabla F(\theta_g^t),g_g^t\rangle]\right]
\le -\mu\eta_t(F(\theta_g^t)-f^*) - \\ 
\frac{\eta_t}{2c}\left\|\sum\limits_{j=1}^{c}\nabla F_j(\theta_j^t)\right\|^2
 +\frac{\eta_t L^2}{2c}\sum\limits_{j=1}^c \|\theta^t_g-\theta^t_j\|^2.
\end{array}
\]
\end{corollary}
Again, we have a slight modification of the result~\cite[Lemma B.5]{haddadpour2019convergence};
\begin{lemma}\label{lem:discrepancy}
\[
\begin{array}{l}
\mathbb{E}\left[\frac{1}{c}\sum\limits_{j=1}^c \|\theta_g^t-\theta_j^t\|^2\right]
\hspace{-1mm}\le \hspace{-1mm}\left(\frac{K+1}{K}\right)\hspace{-1mm} \times \\
\hspace{-1mm}\left(\frac{C_1+\hspace{-.5mm}E}{p} \hspace{-2mm}\sum\limits_{k=t_c+1}^{t-1}\eta_k^2\left[
\frac{\lambda}{p}\left\|\sum\limits_{j=1}^{\bar{c}} \nabla F_j(\theta_j^{(k)})\right\|^2\right.\right. \hspace{-3mm}+\hspace{-1mm}\left.\left.
\sum\limits_{j=\bar{c}+1}^c \|\nabla F_j(\theta_j^{k})\|^2\right]\right) \\ + \left(\frac{K+1}{K}\right) \times \sum\limits_{k=t_c+1}^{t-1}\frac{\eta_k^2\sigma^2}{B},
\end{array}
\]
\end{lemma}
From which we finally derive the following descending equation,
\begin{equation}\label{eq:recur}
\begin{array}{l}
\mathbb{E}[F(\theta_g^{t+1})]-F^*\le  \Delta_t\mathbb{E}[F(\theta_g^t)-F^*]+c_t + \\
\frac{\eta_t}{2\bar{c}^2}
\left[-1+\lambda L\eta_t \left(\frac{C_1+K}{K}\right)\right]\left\|\sum\limits_{j=1}^{\bar{c}} \nabla F_j(\theta_j^t)\right\|^2 \\
\qquad + B_t\sum\limits_{k=t_c+1}^{t-1} \eta_k^2 \left\|\sum\limits_{j=1}^{\bar{c}} \nabla F_j(\theta^k_j)\right\|^2,
\end{array}
\end{equation}
with $\Delta_t = 1-\mu\eta_t$, $B_t = \frac{\lambda (K+1)\eta_t L^2}{K^2}(C_1+E)$, and 
\begin{equation}
\begin{array}{l}
    c_t = \eta_t \left[\frac{L\sigma^2}{KB}+U_g^2\max\left\{\frac{C_1}{K}+1,C_1+E\right\}\right]\times \\ \qquad \left[\frac{\eta_t}{2}+ \frac{L(K+1)}{K}\sum\limits_{k=t_c+1}^{t-1} \eta_k^2\right],
\end{array}
\end{equation}

and we can follow the same reasoning as in~\cite[Lemma B.7]{haddadpour2019convergence} to see that
\begin{equation}\label{eq:desc}
\mathbb{E}[f(\theta_g^{t+1})]-F^*\le \Delta_t \mathbb{E}\left[F(\theta_g^t)-F^*\right]+c_t
\end{equation}
for sufficiently large $E$ i.e., the form of $c_t$ does not affect the requirements in regards to the constants to ensure the form of $\Delta_t$. The form of $c_t$ does change the final set of constants in the result, which are detailed as:
\begin{equation}\label{eq:finaldetailconv}
\begin{array}{l}
\mathbb{E}\left[F(\theta^t_g)-F^*\right] \le \frac{a^3}{(T+a)^3}
\mathbb{E}\left[F(\theta^0_g)-F^*\right] + \\
\qquad\qquad \left[\frac{4\kappa T(T+2a)}{\mu (T+a)^3}+\frac{256\kappa^2 T(E-1)}{\mu (T+a)^3}\right]\\\qquad\qquad\times\left(\frac{\sigma^2}{KB}+\frac{U_g^2}{L}\max\left\{\frac{C_1}{K}+1,C_1+E\right\}\right).
\end{array}
\end{equation}
Thus, the incorrectly identified clients within the cluster add an additional error term proportional to the square of the gradient bound on these clients.
\subsubsection{Proof of Theorem~\ref{th:convnonconvex}} \label{proof-th2}
We consider summing over the iterations in the bound of Lemma~\ref{lem:discrepancy} to obtain
\[
\begin{array}{l}
\frac{1}{Tc}\sum\limits_{t=0}^{T-1}\sum\limits_{j=1}^c\mathbb{E}[\mathbb{E}_{\mathcal{P}^t}\|\theta^t_g-\theta^t_j\|^2] \le \\
\frac{\lambda\eta^2(2 C_1+E(E+1))}{Tc}\frac{(K+1)}{K} \sum\limits_{t=0}^{T-1} \left\|\sum\limits_{j=1}^{\bar c} \nabla F_j(\theta^t_j)\right\|^2 \\
+\frac{\eta^2(2 C_1+E(E+1))}{Tc}\frac{(K+1)}{K}\sum\limits_{j=\bar c+1}^{c} \|\nabla F_j(\theta^t_j)\|^2 + \frac{\eta^2(K+1)(E+1)\sigma^2}{KB} \\
\le \frac{\lambda\eta^2(2 C_1+E(E+1))}{Tc}\frac{(K+1)}{K} \sum\limits_{t=0}^{T-1} \left\|\sum\limits_{j=1}^{\bar c} \nabla F_j(\theta^t_j)\right\|^2  + \\  \frac{\eta^2(K+1)((E+1)\sigma^2+(C_1+E(E+1))U^2_g)}{KB}.
\end{array}
\]

we can combine this with Lemmas~\ref{lem:desclem} and~\ref{lem:gsqbound} and proceed the same chain of reasoning as in~\cite[Theorem 4.4]{haddadpour2019convergence} to obtain:
\[
\begin{array}{rl}
\frac{1}{T}\sum\limits_{t=0}^{T-1}\left[\mathbb{E}[F(\theta^{t+1}_g)]-F(\theta^t_g)\right] & \le -\frac{1}{T}\sum\limits_{t=0}^{T-1} \frac{\eta}{2} \|\nabla F(\theta^t_g)\|^2 
 \\ & + \frac{L\eta^2}{2}\left(\frac{\sigma^2}{KB}+\left(\frac{c}{K}+1\right)U_g^2\right)  \\ 
&
+ \frac{\eta^3 L^2(K+1)((E+1)\sigma^2)}{KB} \\ & + \frac{\eta^3 L^2(K+1)(C_1+E(E+1)U^2_g)}{KB}
\end{array}
\]
from which the final result follows.


\section{Consistency with other Distribution Similarity Metrics}
\label{toy example}

We provide a simple synthetic example to show that our proposed method can effectively capture the similarity/dissimilarity between distributions by leveraging  the principal angle between of the data subspaces spanned by the first $p$ significant left singular vectors of the data. In particular, as shown in Table~\ref{t3}, in each column we sample 100 data points randomly from two Multivariate Gaussian Distribution of dimension 20 with different ${\mathbf{\mu}}$, and ${\mathbf{\Sigma}}$. We then compare our approach with the widely used distance measures including Bhattacharyya distance~\cite{kailath1967divergence},   Maximum Mean Discrepancy (MMD)~\cite{MMD2012}, and Kullback–Leibler (KL) distance~\cite{KL-Gaussian} in inspecting the similarity/dissimilarity of the two distributions. Taking the first two columns into account, by fixing ${\mathbf{\Sigma}}$ and changing ${\mathbf{\mu}}$, we are increasing the distance between the distributions and we expect to see a bigger distance in the second column compared to the first column. A similar discussion can be made about the comparison of the third and fourth columns.


\begin{table}[htbp]
 		\caption{Demonstrating that our proposed method can capture the similarity/dissimilarity between two distributions and its results are consistent with that of the well-known distance measures. The results of PACFL is presented as $x (y)$, where $x$ obtained from Eq.~\ref{adj1}, and $y$ obtained from Eq.~\ref{adj2}. We let $p$ of $U_p$ be 3.}

	\centering
	\resizebox{\columnwidth}{!}{
 	\begin{tabular}{|c||c|c||c|c|}

		\hline
	\makecell{Distribution~/\\Measure} & \makecell{$\mathbf{\mathcal{N}_1}(\bf{\mu}, \bf{\Sigma})$\\$\bf{\mathcal{N}_2}(2\bf{\mu}, \bf{\Sigma})$}  & \makecell{$\bf{\mathcal{N}_1}(\bf{\mu}, \bf{\Sigma})$\\$\bf{\mathcal{N}_2}(3\bf{\mu}, \bf{\Sigma})$}  & \makecell{$\bf{\mathcal{N}_1}(\bf{\mu}, \bf{\Sigma})$\\$\bf{\mathcal{N}_2}(\bf{\mu}, \bf{2\Sigma})$}  
	 & \makecell{$\bf{\mathcal{N}_1}(\bf{\mu}, \bf{\Sigma})$\\$\bf{\mathcal{N}_2}(\bf{\mu}, 5\bf{\Sigma})$}  \\
		
\hline\hline
	
Bhattacharyya~\cite{kailath1967divergence}& 3.07 & 10.31 & 1.14 & 2.13\\
		\hline
 KL~\cite{KL-Gaussian}& 15.18 & 47.12 & 3.31 & 5.79\\
		\hline
   MMD~\cite{MMD2012}& 0.6218 & 0.6238 & 0.6234 & 0.6246\\
		\hline
  \rowcolor{LightCyan} \textbf{PACFL}& 10.73 (32.2) & 18.41 (55.24) & 8.93 (26.79) & 13.72 (41.15)\\
		\hline
		
	\end{tabular}
}
	\label{t3}
\end{table}

\section{Background and Related Work}
\label{ Background and Related Work}
\subsection{Global Federated Learning with Non-IID Data}

The purpose of global federated learning is to train a single global model that minimizes the empirical risk function over the union of the data across all clients. FedAvg~\cite{mcmahan2017communication} simply performs parameter averaging over the client models after several local SGD updates and produces a global model. However, it has been empirically shown that under Non-IID settings, this approach is challenging due to weight divergence (client-drift) and unguaranteed convergence~\cite{zhao2018-non-iid, fedprox-smith-2020, haddadpour2019convergence}. 
To alleviate the model drift, \cite{zhao2018-non-iid} proposed to share a small subset of data across all clients with Non-IID data.
FedProx~\cite{fedprox-smith-2020} incorporates a proximal term to the local training objective to keep the models close to the global model. SCAFFOLD~\cite{scaffold-2020} models data heterogeneity as a source of variance among clients and employs a variance reduction technique. It estimates the update direction of the global model and that of each client. Then, the drift of local training is measured by the difference between these two update directions. Finally, SCAFFOLD corrects the local updates by adding the drift in the local training. 

Another successor to FedAvg is FedNova~\cite{FedNova-2020}, which takes the number of local training epochs of each client in each round into account to obtain an unbiased global model. It suggests normalizing and scaling the local updates of each party according to their number of local epochs before updating the global model. While these advances may make a global model more robust under local data heterogeneity, they do not directly address local-level data distribution performance relevant to individual clients. Further, they yield poor results in practice due to
the presence of Non-IID data where only a subset of all potential features are useful to each client. Therefore, in highly personalized scenarios, the target local model may be fairly different from the global aggregate~\cite{Sreeram2019}. We note that there are a couple of  works~\cite{distributedSVD, distributed-PCA-2022} that are very different from the regular FL works. They are about computing a global truncated SVD or PCA in a distributed manner, respectively.


\subsection{Federated Learning from a Client's Perspective }

To deal with the aforementioned challenges under Non-IID settings, other approaches aim to build a personalized model for each client by customizing the global model via local fine-tuning~\cite{Mansour-federated-2020, fallah2020personalized, liang2020think}. In these schemes, the global model serves as a start point for learning
a personalized model on the local data of every client. However, this global model would not serve as a good initialization if the underlying data distribution of clients are substantially different from each other. Similarly, multi-task learning aims to train personalized models for multiple related tasks in a centralized manner by learning the relationships between the tasks from data~\cite{smith2017federated, Multi-task2016}.


\subsection{Clustered Federated Learning}

Alternatively, clustered FL approaches~\cite{Ghosh-federated-2020, sattler-clustered-fl-2021, Mansour-federated-2020} aim to
cluster clients with similar local data distributions to leverage federated learning per cluster more effectively. However, many rounds of federation may be required until the formation of clusters is stabilized. Specifically, Clustered-FL (CFL)~\cite{sattler-clustered-fl-2021} dynamically forms clusters and assigns client to those clusters based on the cosine similarities of the client updates and the centralized learning. IFCA~\cite{Ghosh-federated-2020} considers a pre-defined number of clusters. At each round, each active client downloads all the available models from the server and selects the best cluster based on the local test accuracy, which is very demanding in terms of communications cost.
In addition, in such a setting where the number of clusters must be fixed \emph{a priori} regardless of the level of data heterogeneity among clients, IFCA could perform poorly
for many clients. This problem can be
quite pronounced under settings with highly skewed Non-IID data, which would require more clusters, or under settings with slightly skewed Non-IID data, which would require fewer clusters. 



\section{ Experimental Results}
\label{Extra Experimental Results}
In this section, we provide details of experimental setup, and  provide additional experiments to conclude the experiments presented in the paper.



\subsection{Non-IID Label Skew}
In Section~\ref{overall-performance} of the  paper we presented the results for Non-IID label skew of $20\%$. Here, we report the results for Non-IID label  skew of $30\%$. Tabel~\ref{tab:niid3} reports the average of final local test accuracy over 100 clients. As is evident, PACFL outperforms all SOTA algorithms by a noticeable margin. 

\footnotesize{
\begin{table}[htbp]
\footnotesize
\begin{center}
\footnotesize
\caption{\small{Test accuracy comparison across different datasets for Non-IID label skew $(30\%)$. For each baseline, the average accuracy over all clients is reported. We run each baseline 3 times for 200 rounds with 10 local epochs and a local batch size of 10.}}
    \color{black}
    \label{tab:niid3}
    \centering
\resizebox{\columnwidth}{!}{
\begin{tabular}{lllll}
            \toprule
            Algorithm & FMNIST & CIFAR-10 & CIFAR-100 & SVHN\\
            \midrule
              SOLO  & $93.93 \pm 0.10$ & $65 \pm 0.65$ & $22.95 \pm 0.81$ & $68.70 \pm 3.13$\\
            FedAvg  & $80.7 \pm 1.9$ & $58.3 \pm 1.2$ & $54.73 \pm 0.41$ & $82.0 \pm 0.7$\\
            FedProx   & $82.5 \pm 1.9$ & $57.1 \pm 1.2$ & $53.31 \pm 0.48$ & $82.1 \pm 1.0$\\
            FedNova   & $78.9 \pm 3.0$ & $54.4 \pm 1.1$ & $54.62 \pm 0.91$ & $80.5 \pm 1.2$\\
            Scafold   & $77.7 \pm 3.8$ & $57.8 \pm 1.4$ & $54.90 \pm 0.42$ & $77.2 \pm 2.0$\\
            LG  & $94.21 \pm 0.40$ & $76.58 \pm 0.16$ & $35.91 \pm 0.20$ & $87.69 \pm 0.77$\\
            PerFedAvg   & $92.87 \pm 2.67$ & $77.67 \pm 0.19$ & $56.42 \pm 0.41$ & $91.25 \pm 1.47$\\
            IFCA  & $95.22 \pm 0.03$ & $80.95 \pm 0.29$ & $67.39 \pm 0.27$ & $93.02 \pm 0.15$\\
            CFL  & $78.44 \pm 0.23$ & $52.57 \pm 3.09$ & $35.23 \pm 2.72$ & $73.97 \pm 4.77$\\
            \rowcolor{LightCyan} \textbf{PACFL}   & $\bf{95.46 \pm 0.06}$ & $\bf{82.77 \pm 0.18}$ & $\bf{67.71 \pm 0.21}$ & $\bf{93.13 \pm 0.27}$ \\
            \midrule
            
        \end{tabular}
    }
\end{center}
\end{table}
}

\normalsize

\subsection{Non-IID Dirichlet Label Skew} \label{supp-dir}

Herein, we distribute the training data between the clients based on the Dirichlet distribution simliar as~\cite{li2021federated}. In particular, we simulated the Non-IID partition into $N$ clients by sampling ${\bf{p}}_i \sim Dir_N(\alpha)$~\footnote{The value of $\alpha$ controls the degree of Non-IID-ness. A big value of $\alpha$ e.g., $\alpha=100$ mimics identical label distribution (IID), while $\alpha=0.1$ results in a split, where the vast majority of data on every client are Non-IID.} and allocating the ${\bf{p}}_{i,j}$ proportion of the training data of class $i$ to client $j$ as in~\cite{li2021federated}. We set the concentration parameter of Dirichlet distribution, $\alpha$ to 0.1 throughout the experiments. Table~\ref{tab:niid-dir} demonstrates the results for Non-IID Dir $(0.1)$.  PACFL has achieved SOTA results over all the reported datasets. The best performance of IFCA with 2 clusters has been used to report the results in Table~\ref{tab:niid-dir}. 


\footnotesize{
\begin{table} [th]
\footnotesize
\begin{center}
\footnotesize
\caption{\small{Test accuracy comparison across different datasets for Non-IID Dir (0.1). For each baseline, the average of final local test accuracy over all clients is reported. We run each baseline 3 times for 200 communication rounds with 10 local epochs and a local batch size of 10. }}

\small
    \color{black}
    \label{tab:niid-dir}
    \centering
\begin{tabular}{lllll}
            \toprule
   
            Algorithm & FMNIST & CIFAR-10 & CIFAR-100 \\ 
            \midrule

             SOLO  &  $69.71 \pm 0.99$ & $41.68 \pm 2.84$ & $16.83 \pm 0.51$ \\ 
            FedAvg  & $82.91 \pm 0.83$ & $38.22 \pm 3.28$ & $44.52 \pm 0.42$ \\ 
            FedProx  &  $84.04 \pm 0.53$ & $42.29 \pm 0.95$ & $45.52 \pm 0.72$ \\ 
            FedNova   & $84.50 \pm 0.66$ & $40.25 \pm 1.46$ & $46.52 \pm 1.34$ \\ 
            Scafold    & $10.0 \pm 0.0$ & $10.0 \pm 0.0$ & $43.73 \pm 0.89$ \\ 
            LG  & $74.96 \pm 1.41$ & $49.65 \pm 0.37$ & $23.59 \pm 0.26$ \\ 
            PerFedAvg  & $80.29 \pm 2.00$ & $53.58 \pm 1.57$ & $33.94 \pm 0.41$ \\ 
            IFCA  & $85.01 \pm 0.30$ & $51.16 \pm 0.49$ & $47.67 \pm 0.28$ \\ 
            CFL  & $74.13 \pm 0.94$ & $42.30 \pm 0.25$ & $31.42 \pm 1.50$ \\ 
            \rowcolor{LightCyan} \textbf{PACFL} & $\bf{85.28 \pm 0.12}$ & $\bf{51.22 \pm 0.25}$ & $\bf{49.80 \pm 0.09}$ \\ 
            
            \midrule
            
        \end{tabular}
\end{center}
\end{table}
}

\normalsize

\subsection{Generalization to Newcomers}
This subsection complements Section~\ref{Generalization-to-Newcomers} of the paper. We show in Algorithm~\ref{alg:newcomers} how PACFL can handle \emph{newcomers} arriving after the federation procedure, to learn their personalized model.




\begin{algorithm}[th]
\footnotesize
\caption{Generalization to newcomers after federation}
\label{alg:newcomers}
\begin{algorithmic}[1]
\STATE \textbf{Server: }{An existing $\mathbf{A}$, $\mathbf{U}$, $\{C_1,...,C_Z \}$, clustering threshold $\beta$}
\STATE \textbf{Require: }{a set of $B$ newcomers and their corresponding first $p$ significant singular vectors, i.e. $\mathbf{U}_{new} = [U_p^1,...,U_p^B]$}
\STATE $\mathbf{A}$, $\mathbf{U}$ = $\rm{PME}(\mathbf{A}$, $\mathbf{U}$, $\mathbf{U}_{new})$ \COMMENT{Alg.~\ref{alg:Func}} 
\STATE $\{C_1,...,C_Z\} \leftarrow \rm{HC}(\mathbf{A}, \beta)$ \COMMENT{Updating the clusters and determining the cluster ID of each new client $k$}
\STATE Each new client $k$ receives the corresponding cluster model $\theta_{g,z} $ from the server 
\STATE ${\rm{FineTune}}(k; \theta_{g,z} )$: by SGD training \COMMENT{this step is optional} 
\end{algorithmic}
\end{algorithm}
\vspace{-1mm}

\subsection{Communication Cost}
This subsection complements Section~\ref{comm-cost-section} of the paper.

\begin{figure*}[htbp]
  \begin{minipage}{\textwidth}
    \centering
   \includegraphics[width=.188\pdfpagewidth]{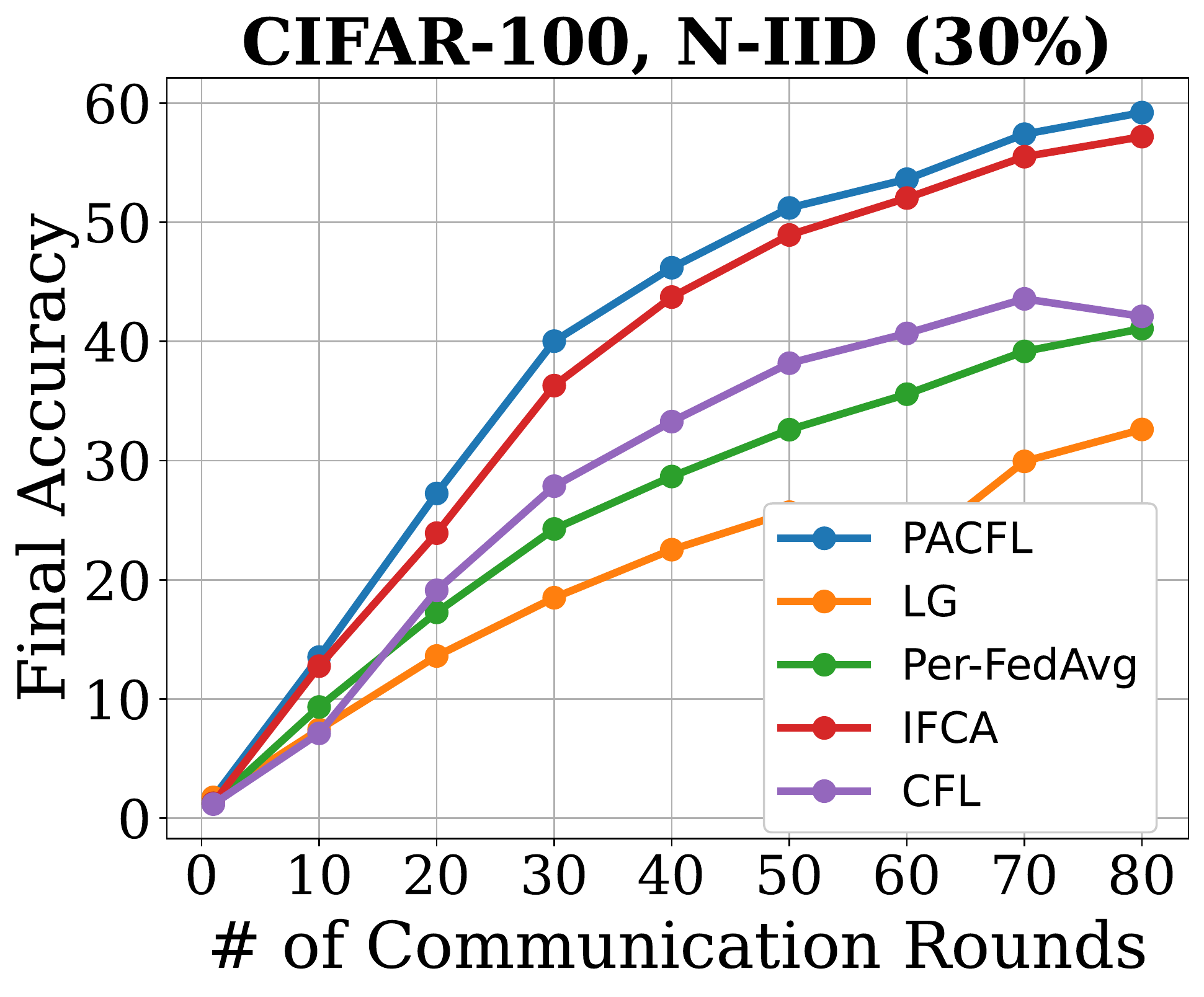}\quad
   \includegraphics[width=.188\pdfpagewidth]{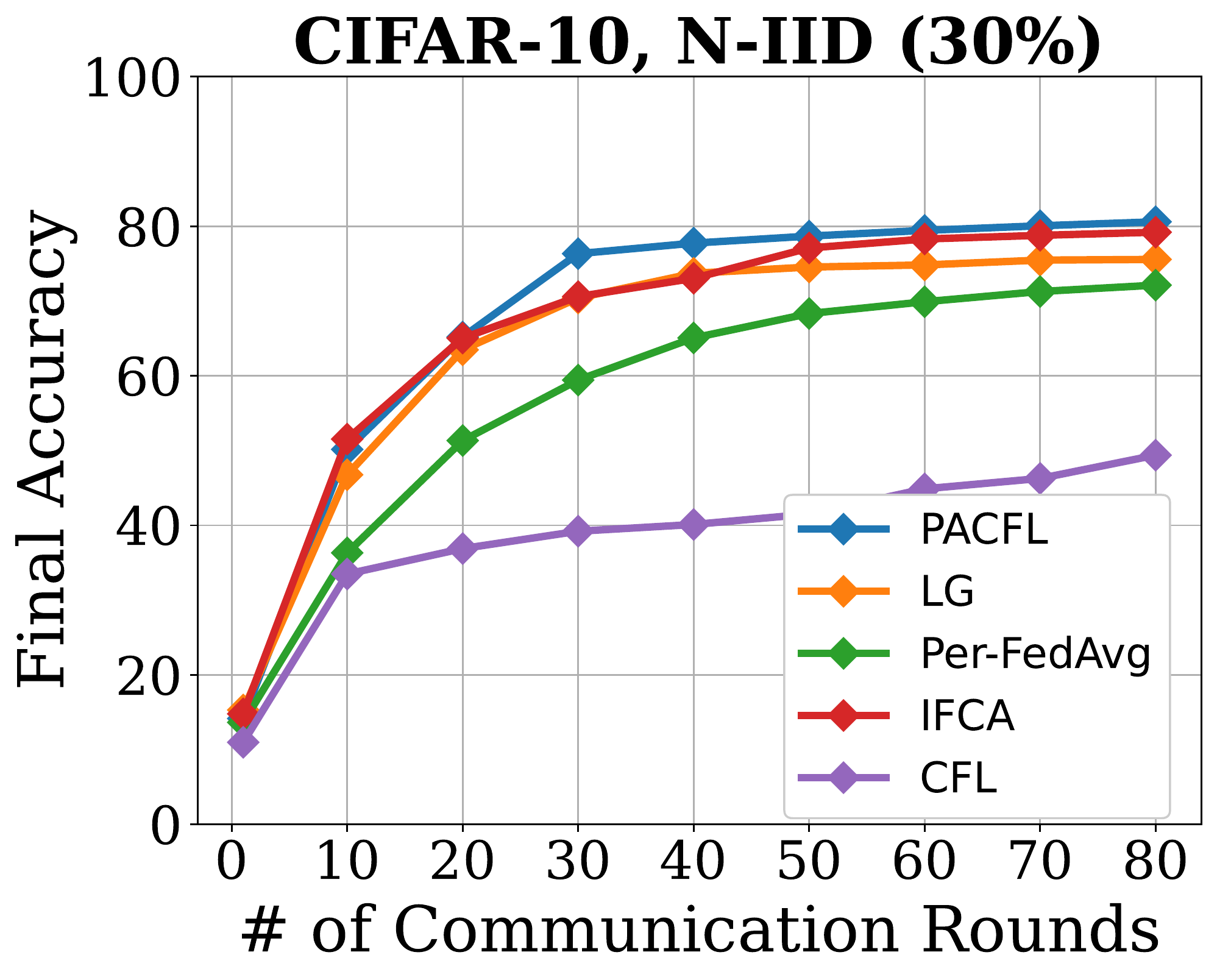}\quad
    \includegraphics[width=.193\pdfpagewidth]{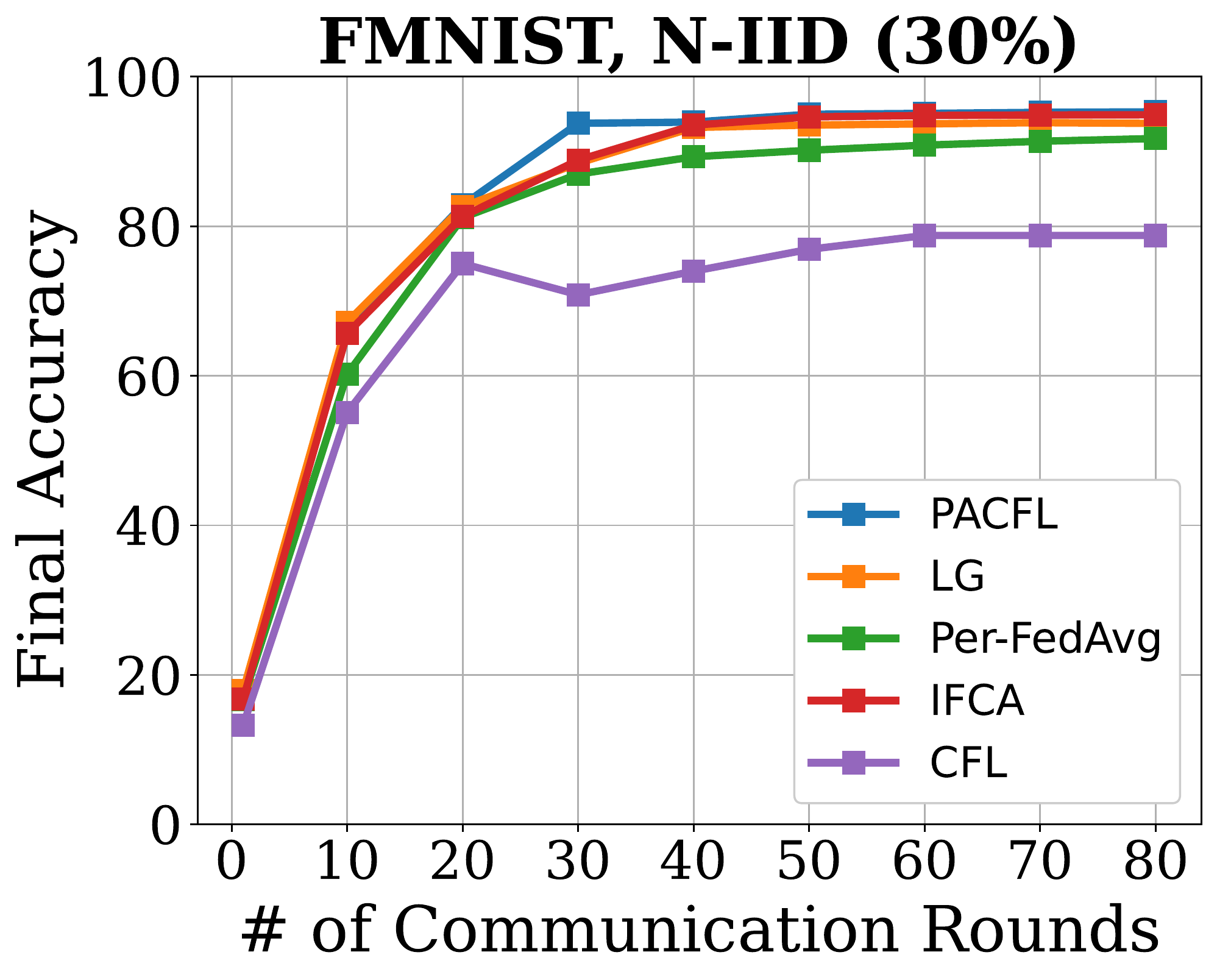}\quad
    \includegraphics[width=.19\pdfpagewidth]{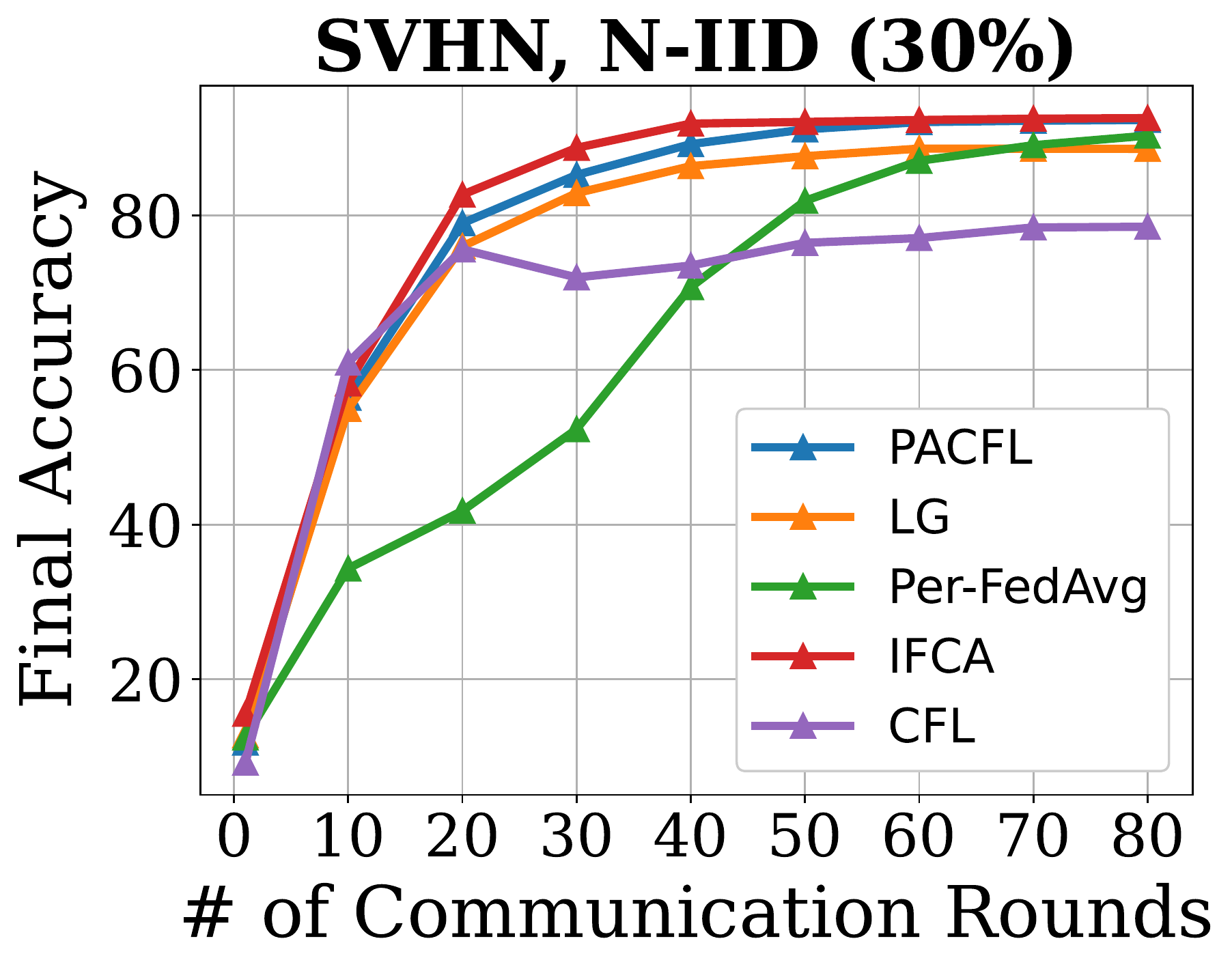}
 \caption{Test accuracy versus number of communication rounds for Non-IID ($30\%$). PACFL converges fast to the desired accuracy and consistently outperforms strong competitors, except in SVHN.}
    \label{fig:comm-cost-niid3}
  \end{minipage}\\[1em]
  \vspace{-5mm}
\end{figure*}

We consider a
limited communication rounds budget of $80$ for all personalized baselines and present the average
of final local test accuracy over all clients versus number of communication rounds for Non-IID
label skew $(30\%)$ in Fig.~\ref{fig:comm-cost-niid3}. In addition, tables~\ref{tab:comm-3} and~\ref{tab:comm-dir} respectively demonstrate the communication cost (in {\bf{Mb}}) required to achieve the
target test accuracies for Non-IID label skew (30$\%$), and Non-IID Dir (0.1) across different datasets.


As can be seen, PACFL is drastically more communication-efficient than all the baselines except for LG in all target accuracies. \emph{Global} baselines either cannot reach the target accuracy or require a high communication cost. Focusing on CIFAR-100 in Table~\ref{tab:comm-3}, PACFL can reduce the communication cost by $\times(1.6-3.2)$ to reach the target accuracies. The communication cost is particularly crucial, as one single transfer of the parameters of ResNet-9 already takes up 211.88 {\bf{Mb}}. Again considering CIFAR-100 in Table~\ref{tab:comm-3}, to achieve the target accuracy of $50\%$, IFCA requires 3495.19 {\bf{Mb}} of cumulative communication, translating to more than 1.7 orders of magnitude in communication savings for PACFL. This is because in every communication round, all clusters (models) at the server will have to be sent to all the participating clients which significantly increases the communication cost. 
Tables~\ref{tab:comm-20},~\ref{tab:comm-3} and~\ref{tab:comm-dir}
confirm that PACFL requires fewer communication cost/rounds compared to the SOTA algorithms to reach the target accuracies.



\begin{table}[htbp]
    \color{black}
\caption{Comparing different FL approaches for Non-IID label skew (30$\%$) in terms of the required communication cost in {\bf{Mb}} to reach target top-1 average local test accuracy. We evaluate on FMNIST, CIFAR-10, CIFAR-100, and SVHN.}    \label{tab:comm-3}
    \centering
    {\footnotesize
        \begin{tabular}{p{10mm}p{15mm}p{15mm}p{15mm}p{8mm}}
            \toprule
            Algorithm & FMNIST & CIFAR-10 & CIFAR-100 & SVHN\\
            \midrule
           Target &  $80\%$ & $70\%$ & $50\%$ & $80\%$ \\ 
            \midrule
            FedAvg  & $79.36$ & $--$ & $4237.37$ & $71.43$\\
            FedProx   & $71.43$ & $--$ & $4237.37$ & $71.43$\\
            FedNova   & $--$ & $--$ & $3601.98$ & $79.36$\\
            Scafold    & $--$ & $--$ & $3305.11$ & $--$\\
            LG  & $\bf{1.26}$ & $\bf{2.11}$ & $--$ & $\bf{1.76}$\\
            PerFedAvg   & $7.54$ & $23.81$ & $6356.06$ & $18.65$\\
            IFCA  & $11.30$ & $16.66$ & $3495.19$ & $10.71$\\
            CFL  & $--$ & $--$ & $--$ & $--$\\
            \rowcolor{LightCyan} \textbf{PACFL}  & $7.53$ & $10.31$ & $\bf{1991.60}$ & $8.73$\\
            \bottomrule
        \end{tabular}
    }
\end{table}

\begin{table}[htbp]
\vspace{-3mm}
    \color{black}
\caption{Comparing different FL approaches for Non-IID Dir (0.1) in terms of the required communication cost in {\bf{Mb}} to reach target top-1 average local test accuracy. We evaluate on FMNIST, CIFAR-10, and CIFAR-100.} 

\label{tab:comm-dir}
    \centering
    {\footnotesize
        \begin{tabular}{p{10mm}p{15mm}p{15mm}p{15mm}}
            \toprule
            Algorithm & FMNIST & CIFAR-10 & CIFAR-100 \\ 
            \midrule
    Target &  $75\%$ & $45\%$ & $40\%$ \\ 
            \midrule
            
            FedAvg  & $9.92$ & $--$ & $3389.47$ \\ 
            FedProx   & $9.92$ & $--$ & $2965.52$ \\ 
            FedNova   & $9.92$ & $--$ & $3178.03$ \\ 
            Scafold    & $--$ & $--$ & $5402.44$ \\ 
            LG   & $14.09$ & $\bf{2.46}$ & $--$ \\ 
            PerFedAvg   & $31.74$ & $18.25$ & $--$ \\ 
            IFCA  & $13.09$ & $13.09$ & $4575.89$ \\ 
            CFL   & $79.36$ & $--$ & $--$ \\ 
            \rowcolor{LightCyan} \textbf{PACFL}  & $\bf{8.73}$ & $8.73$ & $\bf{2372.84}$ \\ 
            \bottomrule
        \end{tabular}
    }
\end{table}

\normalsize
\subsection{Globalization vs Personalization Trade-off}

\begin{figure*}[thbp]
  \begin{minipage}{\textwidth}
    \centering
   \includegraphics[width=.188\pdfpagewidth]{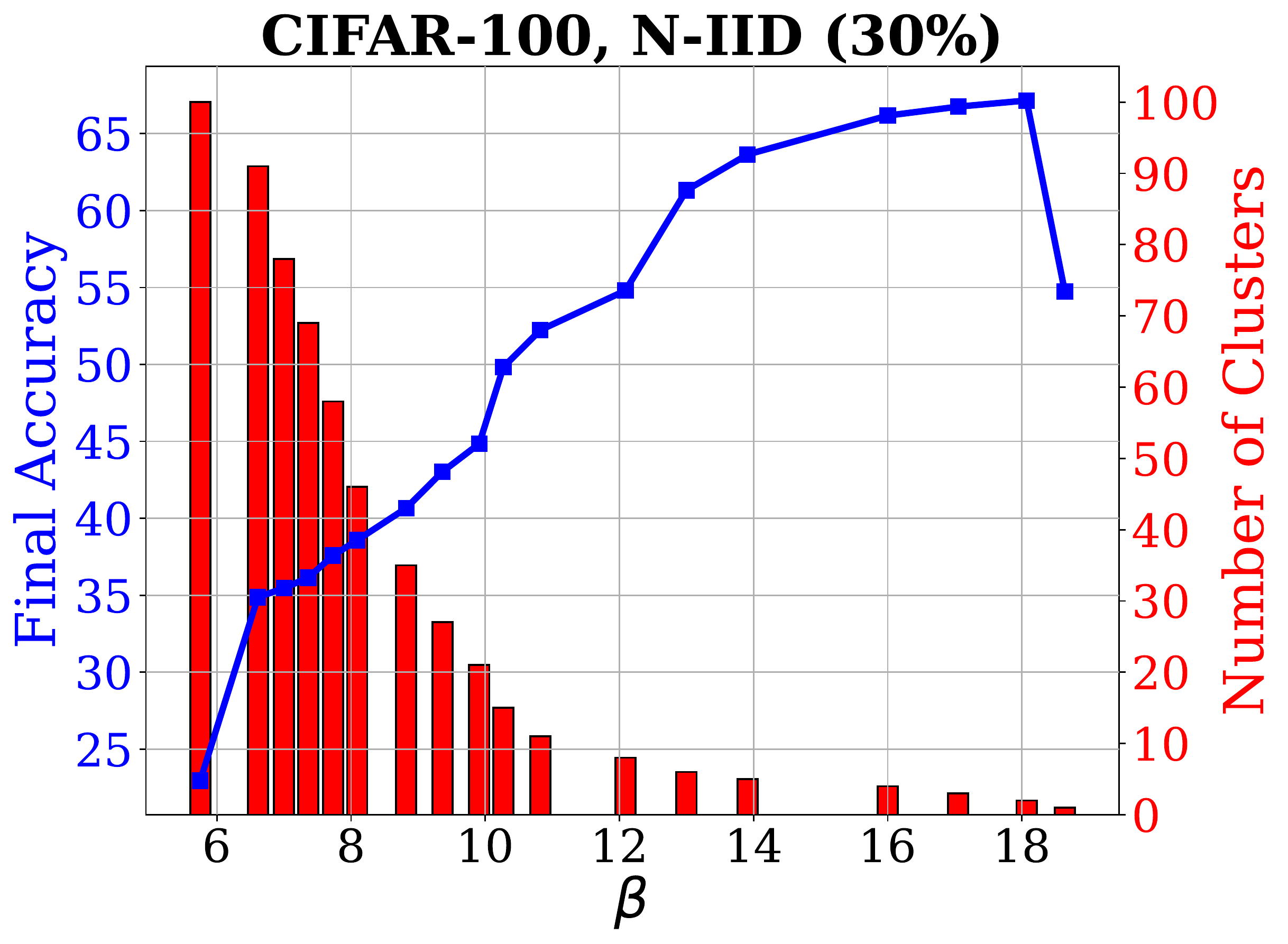}\quad
   \includegraphics[width=.188\pdfpagewidth]{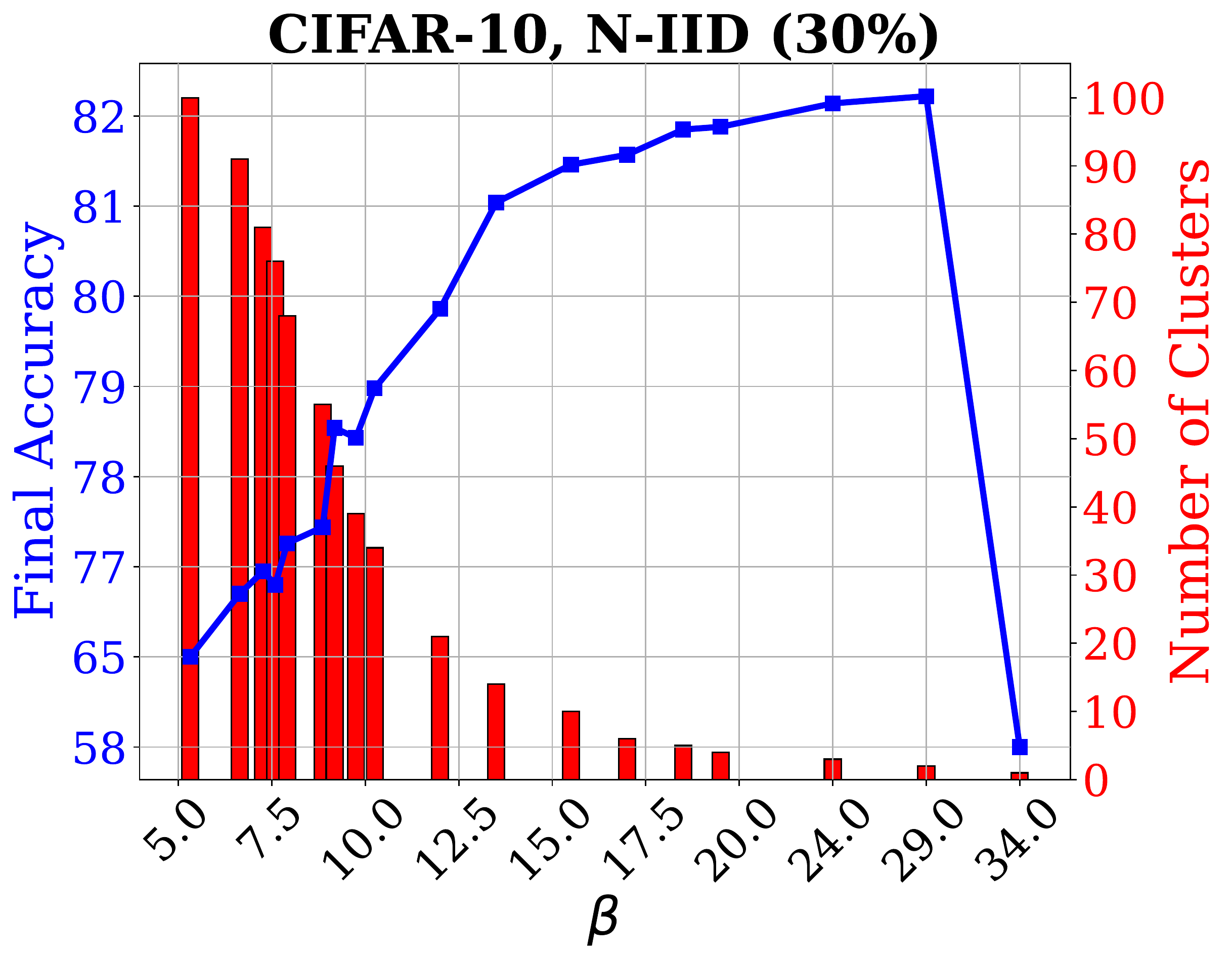}\quad
    \includegraphics[width=.193\pdfpagewidth]{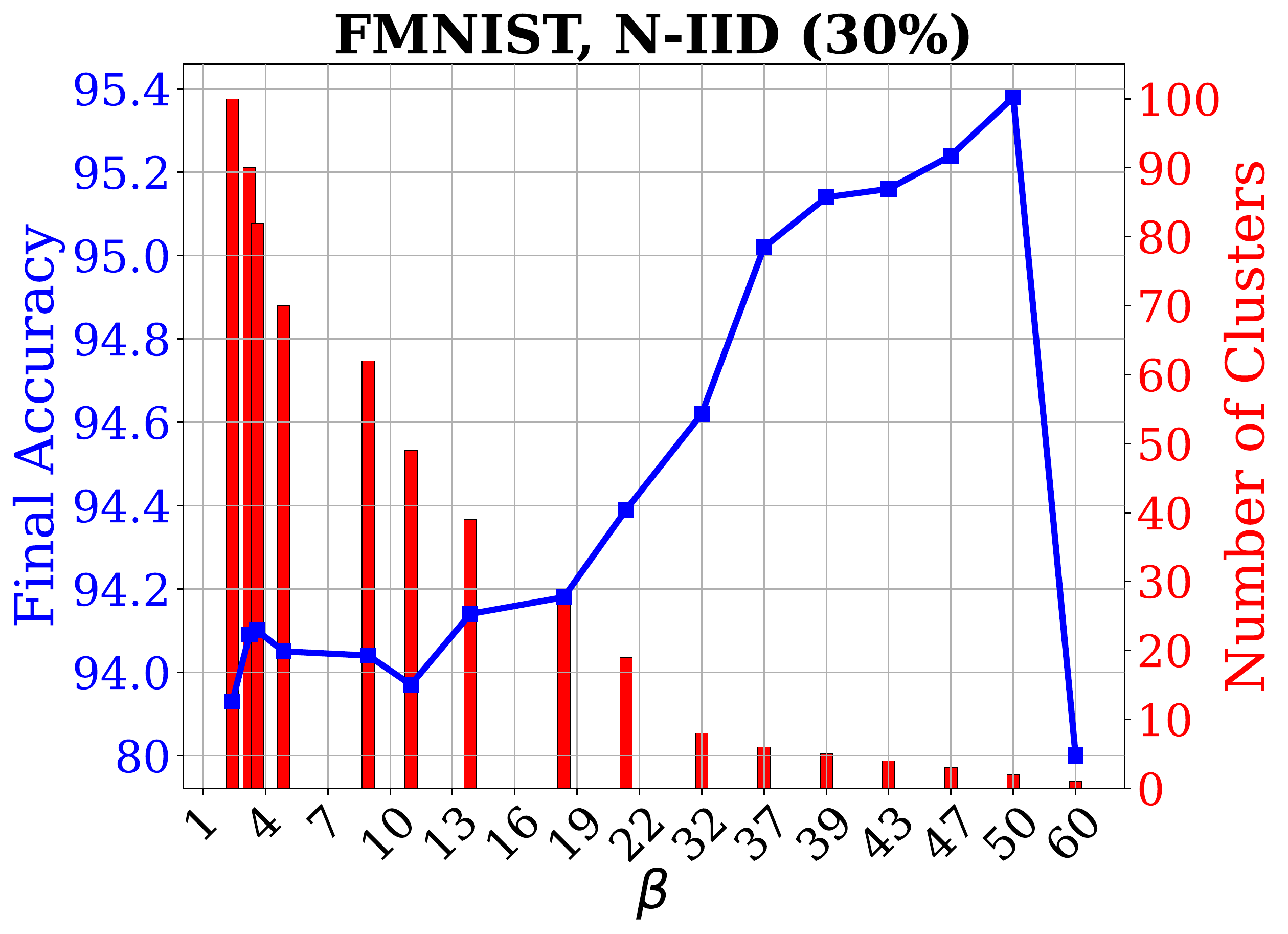}\quad
    \includegraphics[width=.19\pdfpagewidth]{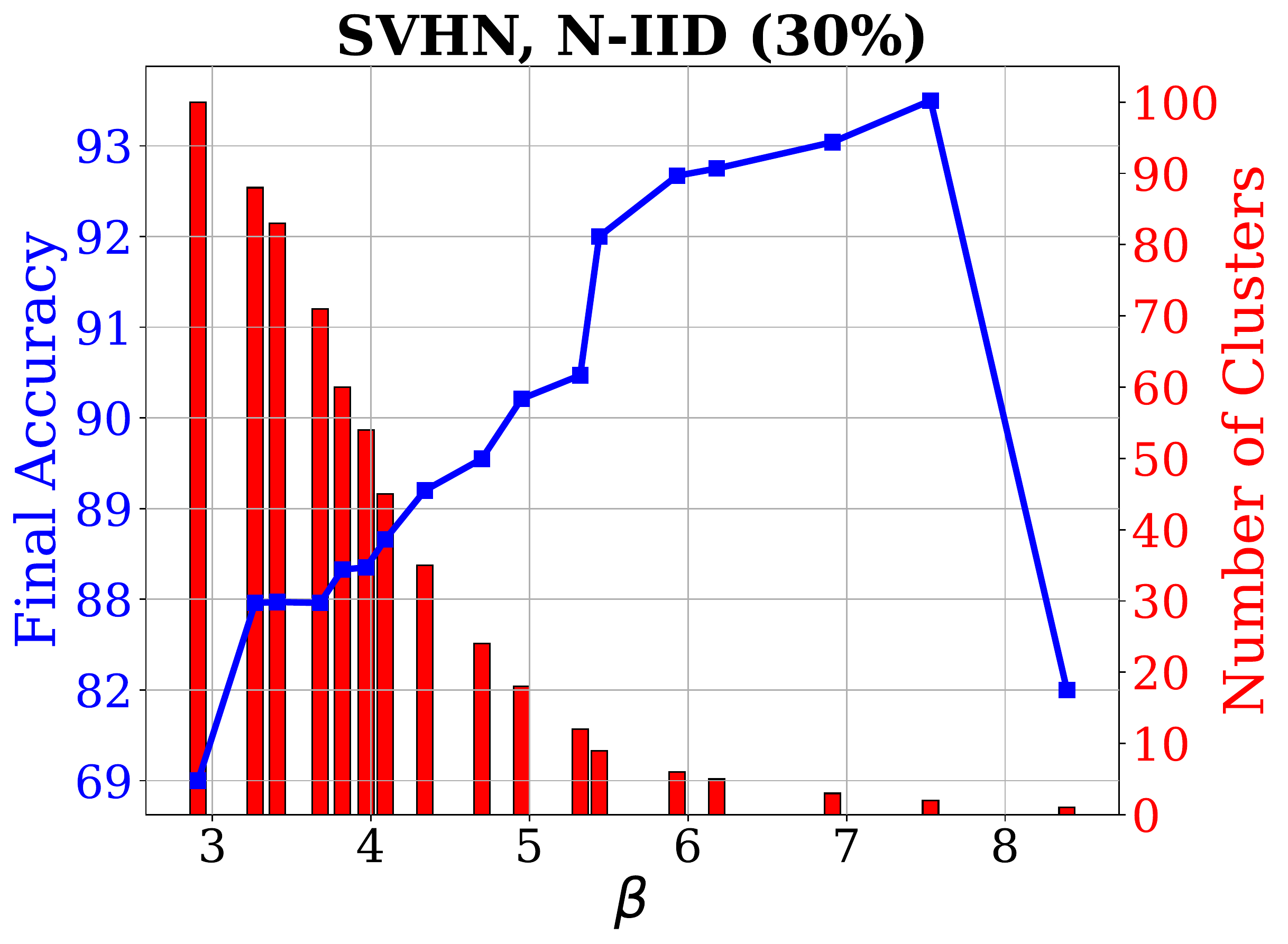}

\vspace{-4mm}
 \caption{Test accuracy performance of PACFL versus the clustering threshold $\beta$ (when the proximity matrix obtained  as in Eq.~\ref{adj2}), and the number of fitting clusters for Non-IID label skew ($30\%$) on CIFAR-10/100, FMNIST, and SVHN datasets. Each point in the plots are obtained by 200 communication rounds with local epoch of 10, local batch size of 10 and SGD local optimizer. This figure complements Fig.~\ref{fig:beta20} in the main paper.}
    \label{fig:beta30}
  \end{minipage}\\[1em]
\end{figure*}
\label{global-personal-trade-off}

This section complements section~\ref{Global-personal-trade-off} of the paper. In this section, we present further experimental results showing the globalization vs personalization trade-off for Non-IID label skew (30$\%$). Fig.~\ref{fig:beta30}, which complements Fig.~\ref{fig:beta20} of the paper, visualizes the accuracy performance behavior of PACFL versus different values of $\beta$. We can see the same behaviour as in Fig.~\ref{fig:beta20} and again CIFAR-10/100, SVHN, FMNIST datasets benefit from a high degree of globalization under Non-IID label skew (30$\%$).



\normalsize
\normalsize

\subsection{Clustering Analysis}

\begin{figure}[thbp]
\vspace{-1.1cm}
    \centering     
   \includegraphics[width=1\columnwidth,height=4.2cm]{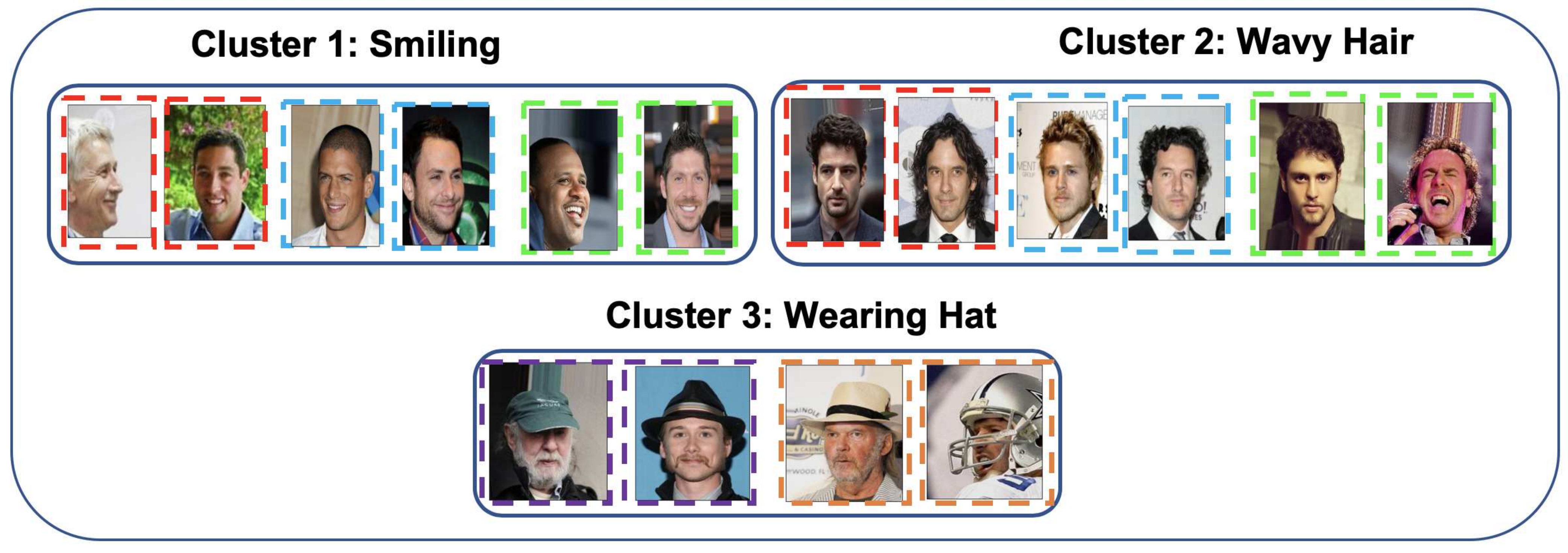}
   \vspace{-0.72cm}
   \caption{\small{A case study on a setting consisting of three clusters that extracted from a random round of federation to testify whether the clients populated into one cluster have similar attribute/data distribution. The images with the same colored edge in each cluster belong to a particular client participated in that round. }}
   \vspace{-0.5cm}
   \label{fig:contribution-celebA}
\end{figure}


To further evaluate the effectiveness of our proposed approach, i.e., leveraging the principal angle between the subspaces spanned by the first $p$ significant left singular vectors of the clients' data and whether the clients populated into one cluster have similar data/attributes/distributions, we conducted a clustering analysis via an illustration. Consider the scenario where we are interested in forming clusters of images for a face recognition task on the CelebA~\cite{liu2015faceattributes} dataset. In particular, assume that we are inclined to form clusters of clients for each attribute and find a good model for each cluster. To do so, we partition the data of CelebA based on three attributes namely ``smiling", ``wavy hair", and ``wearing hat" such that each partition only has one of the attributes and none of the partitions have overlapped with the other partitions on the selected attributes. Then, we divide each of the partitions into shards and randomly assign to all clients and run some rounds of federation. To intuitively judge whether the clients populated into the same cluster have similar data/attributes, we monitored the clustering result of of local models derived from PACFL on the CelebA dataset. As shown in Fig.~\ref{fig:contribution-celebA}, the face recognition task in cluster~1, cluster~2, cluster~3, is likely to handle the Smiling faces, celebrities with wavy hair, and the people with wearing hats.

\normalsize
\section{Implementation} \label{imp}
\textbf{We have released our implemented code for PACFL at 
~\url{https://github.com/MMorafah/PACFL}}.
To be consistent, we adapt the official public codebase of Qinbin et al.~\cite{li2021federated} \footnote{\label{qinbin-imp} \url{https://github.com/Xtra-Computing/NIID-Bench}} to implement our proposed method and all the other baselines with the same codebase in PyTorch V. 1.9. We used the public codebase of LG-FedAvg~\cite{liang2020think}\footnote{\label{lg-imp}~\url{https://github.com/pliang279/LG-FedAvg}}, Per-FedAvg~\cite{fallah2020personalized} \footnote{~\url{https://github.com/CharlieDinh/pFedMe}}, Clustered-FL (CFL) ~\cite{sattler-clustered-fl-2021} \footnote{\label{cfl-imp}~\url{https://github.com/felisat/clustered-federated-learning}}, and IFCA~\cite{Ghosh-federated-2020}  \footnote{~\url{https://github.com/jichan3751/ifca}} in our implementation. For all the global benchmarks, including FedAvg~\cite{mcmahan2017communication}, FedProx~\cite{fedprox-smith-2020}, FedNova~\cite{FedNova-2020}, Scaffold~\cite{scaffold-2020} we used the official public codebase of Qinbin et al.~\cite{li2021federated}~\footref{qinbin-imp}. We run the experiments on RCI clusters~\footnote{The access to the computational infrastructure of the OP VVV funded project CZ.02.1.01/0.0/0.0/16\_019/0000765 ``Research Center for Informatics'' is also gratefully acknowledged for running the experiments.}.

Unlike the original paper and the official implementation of LG-FedAvg\cite{liang2020think}~\footref{lg-imp}, we did not use the model found by performing many rounds of FedAvg as the initialization for LG-FedAvg in our implementation. Instead, for a fair comparison we initialized the models randomly with the same random seed in all other baselines. Also, the original implementation of LG-FedAvg~\footref{lg-imp} computes the average of the maximum local test accuracy for each client over the entire federation, but we compute and report the average of final local test accuracy of all clients for a fair comparison. In all the experiments, we have reported the results over all baselines on the same 100 clients with the same Non-IID partitions to have a fair comparison.

\subsection{Implementation Details for MIX-4} \label{mix4-details}
As stated in Section~\ref{main-mix4} of the paper, we set the number of clients to 100 and distribute CIFAR-10, SVHN, FMNIST, USPS amongst 31, 25, 27, 14 clients such that each client receives 500 samples from all the available classes in the corresponding dataset (50 samples per each class). We further zero-pad FMNIST, and USPS images to make them $32 \times 32$, and repeat them to have 3 channels. This pre-processing for FMNIST, and USPS is required to make the images the same size as CIFAR-10 and SVHN so that we can have a consistent model architecture in this task. We used LeNet-5 architecture with the details in Table~\ref{tab:lenet5}, and modified the last layer to have 40 outputs corresponding to the 40 different labels in total (each dataset has 10 classes). The number of communication rounds is set to 50, and each client performs 5 local epochs. We run each baseline 3 times and report mean and standard deviation for average of local test accuracy in Table~\ref{tab:mix4} of the main paper. Tables~\ref{tab:param1}, and ~\ref{tab:param2} present more details about other hyper-parameter grids used in this experiment.


\subsection{Hyper-parameters \& Architectures} \label{hyper-parameters}

Tables \ref{tab:param1} and \ref{tab:param2} summarize the hyper-parameter grids used in our experiments. In all experiments, we use SGD as the local optimizer and the local batch size is 10. For the hyper-parameters exclusive for certain approaches, e.g. number of clusters in CFL or IFCA, we used the same values as described in the original papers.

For CFL, we lower the learning rate to address its divergence problem on some of the datasets as also specified in the CFL paper~\cite{sattler-clustered-fl-2021}. Moreover, the CFL implementation by the authors\footref{cfl-imp} does not include the parameter $\gamma$ despite of the large body of discussion in the paper, so we leave it out as well. 


\normalsize
{
\begin{table}[ht]
\vspace{-0.1cm}
\footnotesize
\centering
\caption{\footnotesize The details of LeNet-5 architecture used for the FMNIST, SVHN, CIFAR-10, and Mix-4 datasets.}
{
\resizebox{0.8\columnwidth}{!}{
\begin{tabular}{l|l}
\toprule
\multicolumn{1}{l|}{\textbf{Layer}} & \multicolumn{1}{c}{\textbf{Details}} \\
\midrule
\multirow{3}{*}{layer 1} & Conv2d(i=3, o=6, k=(5, 5), s=(1, 1)) \\
                         & ReLU() \\
                         & MaxPool2d(k=(2, 2)) \\
\midrule
\multirow{3}{*}{layer 2} & Conv2d(i=6, o=16, k=(5, 5), s=(1, 1)) \\
                         & ReLU() \\
                         & MaxPool2d(k=(2, 2)) \\
\midrule
\multirow{2}{*}{layer 3} & Linear(i=400 (256 for FMNIST), o=120) \\
                         & ReLU() \\
\midrule
\multirow{2}{*}{layer 4} & Linear(i=120, o=84) \\
                         & ReLU() \\
\midrule
layer 5    & Linear(i=84, o=10 (100 for CIFAR-100, and 40 for Mix-4)) \\
\bottomrule
\end{tabular}
}
}
\label{tab:lenet5}
\end{table}
}

For IFCA, we observe that  the results with 2, and 3 clusters 
are almost the same and higher than 3 clusters does not improve the results as mentioned in the original paper~\cite{Ghosh-federated-2020}. Hence, for a fair comparison we used 2 clusters (the best performance of IFCA) in all of our experiments including the results of Tables~\ref{tab:niid2},~\ref{tab:niid3}, and ~\ref{tab:niid-dir}, except in~\ref{tab:mix4} where we used both $2$ and $4$ clusters. 

For PACFL, we used $p=3$ for reporting the results in Tables~\ref{tab:niid2}, ~\ref{tab:niid3}, and ~\ref{tab:mix4}. For the results reported in Table~\ref{tab:niid-dir}, we used $p=5$.

We also observed that in most cases the optimal learning rate for FedAvg was also the optimal one for the other baselines, except for Scaffold which had lower learning rates. Tables~\ref{tab:lenet5} and~\ref{tab:resnet9} provide the details of the used model architectures, where $i$ stands for the number of input channels of conv layers and the number of input features of fc layers; $o$ stands for the number of output channels of conv layers and the number of output features of fc layers; $k$ stands for kernel size; $s$ stands for stride, finally, $g$ represents the number of groups.


\begin{table}[h]
\vspace{-2mm}
\footnotesize
\centering
\caption{\footnotesize The details of ResNet-9 architecture used for the CIFAR-100 dataset.}
\resizebox{0.8\columnwidth}{!}{

\begin{tabular}{l|l|l}
\toprule
\multicolumn{1}{l|}{\textbf{Block}} & \multicolumn{1}{c|}{\textbf{Details}} & \multicolumn{1}{l}{\textbf{Input}} \\
\midrule
\multirow{3}{*}{block 1}    & Conv2d(i=3, o=64, k=(3, 3), s=(1, 1))    & \multirow{3}{*}{image} \\
                            & GroupNorm(g=32, o=64)                    & \\
                            & ReLU()                                   & \\
\midrule
\multirow{4}{*}{block 2}    & Conv2d(i=64, o=128, k=(3, 3), s=(1, 1))  & \multirow{4}{*}{block 1} \\
                            & GroupNorm(g=32, o=128)                   & \\
                            & ReLU()                                   & \\
                            & MaxPool2d(k=(2, 2))                      & \\
\midrule
\multirow{6}{*}{block 3}    & Conv2d(i=128, o=128, k=(3, 3), s=(1, 1)) & \multirow{6}{*}{block 2} \\
                            & GroupNorm(g=32, o=128)                   & \\
                            & ReLU()                                   & \\
                            & Conv2d(i=128, o=128, k=(3, 3), s=(1, 1)) & \\
                            & GroupNorm(g=32, o=128)                   & \\
                            & ReLU() \\
\midrule
\multirow{4}{*}{block 4}    & Conv2d(i=128, o=256, k=(3, 3), s=(1, 1)) & \\ 
                            & GroupNorm(g=32, o=256)                   & block 2 + \\
                            & ReLU()                                   & block 3 \\
                            & MaxPool2d(k=(2, 2))                      & \\
\midrule
\multirow{4}{*}{block 5}    & Conv2d(i=256, o=512, k=(3, 3), s=(1, 1)) & \multirow{4}{*}{block 4} \\
                            & GroupNorm(g=32, o=512)                   & \\
                            & ReLU()                                   & \\
                            & MaxPool2d(k=(2, 2))                      & \\
\midrule
\multirow{6}{*}{block 6}    & Conv2d(i=512, o=512, k=(3, 3), s=(1, 1)) & \multirow{6}{*}{block 5} \\
                            & GroupNorm(g=32, o=512)                   & \\
                            & ReLU()                                   & \\
                            & Conv2d(i=512, o=512, k=(3, 3), s=(1, 1)) & \\
                            & GroupNorm(g=32, o=512)                   & \\
                            & ReLU() \\
\midrule
\multirow{2}{*}{classifier} & MaxPool2d(k=(4, 4))                      & block 4 + \\ 
                            & Linear(i=512, o=100)                     & block 5\\
\bottomrule
\end{tabular}
}
\label{tab:resnet9}
\end{table}


\onecolumn
\tiny{
\begin{table*}[ht]
\tiny
\caption{The hyper-parameters used for FedAvg, FedProx, FedNova, Scaffold, and SOLO throughout the experiments}
\centering
\resizebox{\textwidth}{!}{
\begin{tabular}{c|l|ccccc}
\toprule
\textbf{Method}           & \multicolumn{1}{c|}{\textbf{Hyper-parameters}} & CIFAR-100 & CIFAR-10  & FMNIST  & SVHN & Mix4 \\
\midrule
\multirow{4}{*}
{\shortstack{
FedAvg}}                & model                                          & ResNet-9  & LeNet-5 & LeNet-5 & LeNet-5 & LeNet-5 \\
                          & learning rate                                  &\{0.1, 0.01, 0.001\} &\{0.1, 0.01, 0.001\} & \{0.1, 0.01, 0.001\} & \{0.1, 0.01, 0.001\} & \{0.1, 0.01, 0.001\} \\
                          & weight decay                                   &{0} &{0} &{0} &{0} &{0}     \\
                          & momentum                                       &{0.9} &{0.9} &{0.9} &{0.9} &{0.9}   \\

\midrule
\multirow{5}{*}
{\shortstack{
FedProx}}                & model                                          & ResNet-9  & LeNet-5 & LeNet-5 & LeNet-5 & LeNet-5 \\
                          & learning rate                                  & \{0.1, 0.01, 0.001\} & \{0.1, 0.01, 0.001\} & \{0.1, 0.01, 0.001\} & \{0.1, 0.01, 0.001\} & \{0.1, 0.01, 0.001\} \\
                          & weight decay                                   &{0} &{0} &{0} &{0} &{0}     \\
                          & momentum                                       &{0.9} &{0.9} &{0.9} &{0.9} &{0.9}   \\
                          & $\mu$                                          & \{0.01, 0.001\} & \{0.01, 0.001\} & \{0.01, 0.001\} & \{0.01, 0.001\} & \{0.01, 0.001\} \\
                          
\midrule
\multirow{4}{*}
{\shortstack{
FedNova}}                & model                                          & ResNet-9  & LeNet-5 & LeNet-5 & LeNet-5 & LeNet-5 \\
                          & learning rate                                  & \{0.1, 0.01, 0.001\} & \{0.1, 0.01, 0.001\} & \{0.1, 0.01, 0.001\} & \{0.1, 0.01, 0.001\} & \{0.1, 0.01, 0.001\} \\
                          & weight decay                                   &{0} &{0} &{0} &{0} &{0}     \\
                          & momentum                                       &{0.9} &{0.9} &{0.9} &{0.9} &{0.9}   \\
\midrule
\multirow{4}{*}
{\shortstack{
Scaffold}}                & model                                          & ResNet-9  & LeNet-5 & LeNet-5 & LeNet-5 & LeNet-5 \\
                          & learning rate                                  & \{0.1, 0.01, 0.001\} & \{0.1, 0.01, 0.001\} & \{0.1, 0.01, 0.001\} & \{0.1, 0.01, 0.001\} & \{0.1, 0.01, 0.001\} \\
                          & weight decay                                   &{0} &{0} &{0} &{0} &{0}     \\
                          & momentum                                       &{0.9} &{0.9} &{0.9} &{0.9} &{0.9}   \\
\midrule
\multirow{4}{*}
{\shortstack{
SOLO}}                & model                                          & ResNet-9  & LeNet-5 & LeNet-5 & LeNet-5 & LeNet-5 \\
                          & learning rate                                  & 0.01 & 0.01 & 0.01 & 0.01 & 0.01 \\
                          & weight decay                                   &{0} &{0} &{0} &{0} &{0}     \\
                          & momentum                                       &{0.5} &{0.5} &{0.5} &{0.5} &{0.5}   \\
\bottomrule
\end{tabular}
}
\label{tab:param1}
\end{table*}
} 


\tiny{
\begin{table}[b]
\vspace{-20cm}
\tiny
\caption{Hyper-parameters used for LG, Per-FedAvg, IFCA, CFL, and our method throughout the experiments}
\centering
\resizebox{\textwidth}{!}{
\begin{tabular}{c|l|ccccc}
\toprule
\textbf{Method}           & \multicolumn{1}{c|}{\textbf{Hyper-parameters}} & CIFAR-100 & CIFAR-10  & FMNIST  & SVHN & Mix4 \\
\midrule
\multirow{6}{*}
{\shortstack{
LG}}                & model                                          & ResNet-9  & LeNet-5 & LeNet-5 & LeNet-5 & LeNet-5 \\
                          & learning rate                                  & 0.01 & 0.01 & 0.01  & 0.01 & 0.01 \\
                          & weight decay                                   &{0} &{0} &{0} &{0} &{0}     \\
                          & momentum                                       &{0.5} &{0.5} &{0.5} &{0.5} &{0.5}   \\
                          & number of local layers                         & 7 & 3 & 3 & 3 & 3 \\
                          & number of global layers                        & 2 & 2 & 2 & 2 & 2 \\
\midrule
\multirow{6}{*}
{\shortstack{
Per-FedAvg}}                & model                                          & ResNet-9  & LeNet-5 & LeNet-5 & LeNet-5 & LeNet-5 \\
                          & learning rate                                  & 0.01  & 0.01 & 0.01 & 0.01 & 0.01  \\
                          & weight decay                                   &{0} &{0} &{0} &{0} &{0}     \\
                          & momentum                                       &{0.5} &{0.5} &{0.5} &{0.5} &{0.5}   \\
                          & $\alpha$                                       &1e-2 &1e-2 &1e-2 &1e-2 &1e-2 \\
                          & $\beta$                                        &1e-3 &1e-3 &1e-3 &1e-3 &1e-3 \\
\midrule
\multirow{5}{*}
{\shortstack{
IFCA}}                & model                                          & ResNet-9  & LeNet-5 & LeNet-5 & LeNet-5 & LeNet-5 \\
                          & learning rate                                  & 0.01 & 0.01 & 0.01 & 0.01 & 0.01 \\
                          & weight decay                                   &{0} &{0} &{0} &{0} &{0}     \\
                          & momentum                                       &{0.5} &{0.5} &{0.5} &{0.5} &{0.5}   \\
                          & number of clusters                                       & 2 & 2 & 2 & 2 & 4 \\
\midrule
\multirow{6}{*}{CFL}     & model                                          & ResNet-9  & LeNet-5 & LeNet-5 & LeNet-5 & LeNet-5\\
                          & learning rate                                  & \{0.1, 0.01, 0.001\} & \{0.1, 0.01, 0.001\} & \{0.1, 0.01, 0.001\} & \{0.1, 0.01, 0.001\} & \{0.1, 0.01, 0.001\} \\
                          & weight decay                                   &{0} &{0} &{0} &{0} &{0}     \\
                          & momentum                                       &{0.5} &{0.5} &{0.5} &{0.5} &{0.5}   \\
                          & $\epsilon_1$                                   &{0.4} &{0.4} &{0.4} &{0.4} &{0.4}   \\
                          & $\epsilon_2$                                   &{1.6} &{1.6} &{1.6} &{1.6} &{1.6}   \\
                          
\midrule
\multirow{5}{*}
{\shortstack{
PACFL}}                & model                                          & ResNet-9  & LeNet-5 & LeNet-5 & LeNet-5 & LeNet-5 \\
                          & learning rate                                  & 0.01 & 0.01 & 0.01 & 0.01 & 0.01 \\
                          & weight decay                                   &{0} &{0} &{0} &{0} &{0}     \\
                          & momentum                                       &{0.5} &{0.5} &{0.5} &{0.5} &{0.5}   \\
                          & number of clusters                             & 2 & 2 & 2 & 2 & 4 \\
                          & number of $\mathbf{U}_p$                       & 3-5 & 3-5 & 3-5 & 3-5 & 3-5 \\

\bottomrule
\end{tabular}
}

\label{tab:param2}
\end{table}
}   

\normalsize

\newpage

\twocolumn

\bibliography{Saeed}

\end{document}